\PassOptionsToPackage{table}{xcolor}
\documentclass[onecolumn]{openhelix}

\usepackage{amsmath,amsfonts,amssymb}
\usepackage{array,tabularx,makecell}
\usepackage{pifont}
\usepackage{float}
\usepackage{bm}
\usepackage{enumitem}
\usepackage{marvosym}
\newcolumntype{Y}{>{\centering\arraybackslash}X}
\newcommand{\Checkmark}{\ding{51}}
\newcommand{\Crossmark}{\ding{55}}
\newcommand{\method}{SLA}
\renewcommand{\title}[1]{\def\titlelist{{\fontsize{15}{18}\selectfont\sffamily\bfseries\textcolor[HTML]{1F4E8C}{#1}}}}
\title{\method{}: Scaling Spatial-Language-Action\\Model via Continual Pre-training}
\author[1,*]{Fuhao Li}
\author[1,*,\dagger]{Wenxuan Song}
\author[1]{Qizhou Huang}
\author[1]{Ziyang Zhou}
\author[1]{Jingbo Wang}
\author[1]{Junwu E}
\author[3]{Pengwei Wang}
\author[4]{Yan Wang}
\author[2]{Shunbo Zhou}
\author[1,\text{\Letter}]{Haoang Li}
\affiliation[1]{The Hong Kong University of Science and Technology (Guangzhou)}
\affiliation[2]{OLA Dimensions}
\affiliation[3]{\mbox{Beijing Academy of Artificial Intelligence}}
\affiliation[4]{Tsinghua University}
\contribution[*]{Equal Contribution}
\contribution[\dagger]{Project Lead}
\contribution[\text{\Letter}]{Corresponding Author}
\abstract{Reliable robotic manipulation requires both semantic understanding and precise control in three-dimensional space. However, vision-language-action (VLA) foundations largely inherit visual representations from 2D vision-language pre-training, while spatial supervision is often introduced only during task-specific adaptation. Developing spatial understanding as a transferable foundation-model capability therefore calls for learning from broader robot experience. We present \method{}, a Spatial-Language-Action Model that scales implicit spatial alignment to continual pre-training and maintains it during downstream post-training. Initialized from a pretrained $\pi_{0.5}$ policy, \method{} jointly optimizes action prediction and alignment between intermediate visual embeddings and spatial representations from a frozen 3D perception model. We curate and unify heterogeneous robot datasets to support spatial learning across diverse embodiments, tasks, and scenes. To accommodate this diversity, we introduce feature-dependent gates that adaptively aggregate visual embeddings from multiple VLA layers for spatial supervision. The teacher further exploits partially available depth and camera calibration alongside RGB observations, enabling effective use of heterogeneous geometric annotations. Probabilistic state dropout encourages reliance on visual cues and mitigates overfitting to demonstration trajectories. All auxiliary components are used only during training, allowing direct reuse of pretrained VLA weights and preserving the original inference architecture. 
Extensive evaluations demonstrate that \method{} achieves state-of-the-art performance on RoboDojo-Real, RoboTwin 2.0 under the Clean2Random setting, RoboCasa-Panda, and LIBERO-Plus, highlighting its strong generalization and downstream data efficiency. Across eight real-world deployment conditions, \method{} further improves the average success rate over $\pi_{0.5}$ by 24.0 percentage points.}
\hypersetup{pdftitle={SLA: Scaling Spatial-Language-Action Model via Continual Pre-training},
  pdfauthor={Fuhao Li, Wenxuan Song, Qizhou Huang, Ziyang Zhou, Jingbo Wang, Junwu E, Pengwei Wang, Yan Wang, Shunbo Zhou, Haoang Li}}
\newcommand{\githubicon}{\raisebox{-1.5pt}{\includegraphics[height=1.05em]{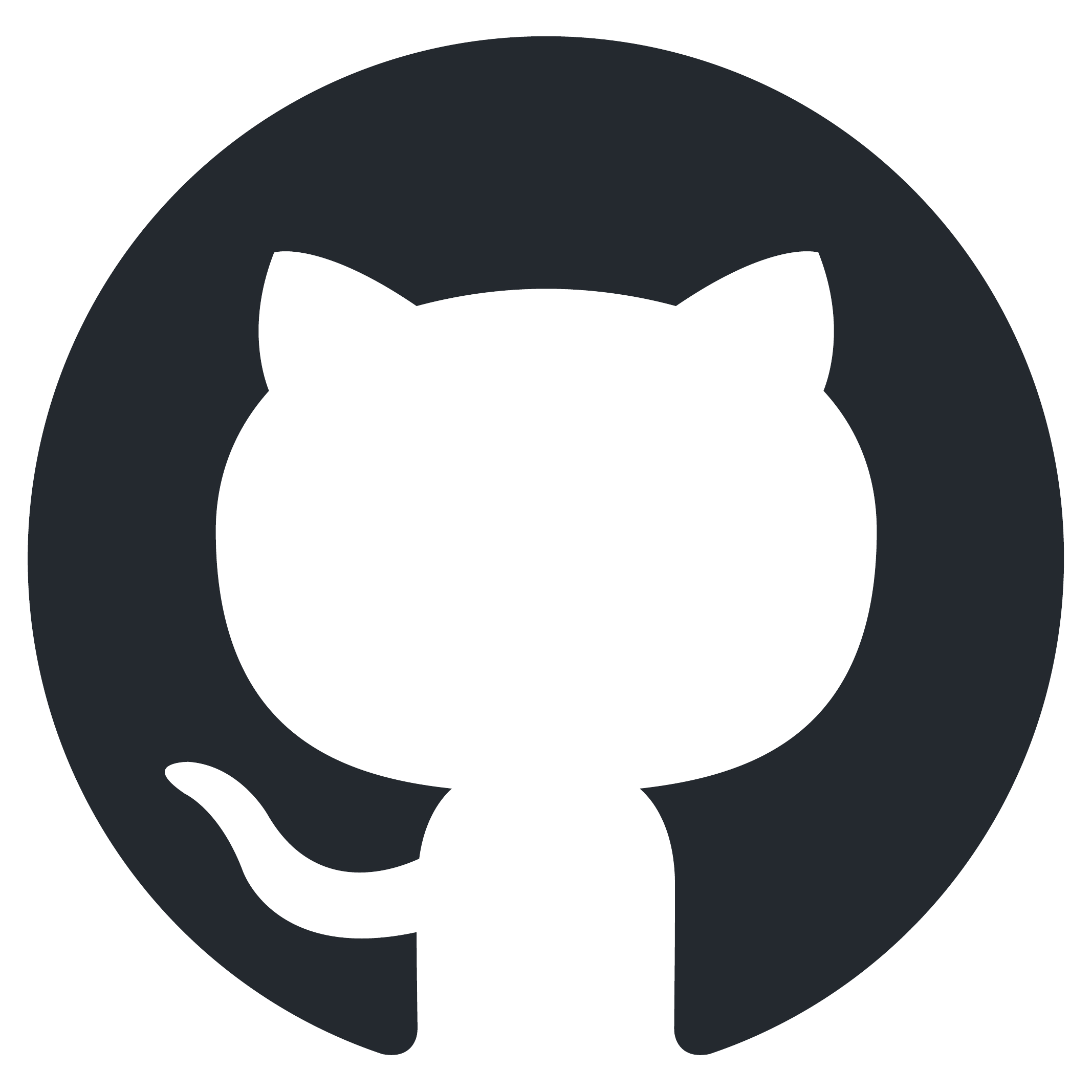}}}
\newcommand{\projecticon}{\raisebox{-1.5pt}{\includegraphics[height=1.05em]{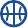}}}
\metadata[Code]{\href{https://github.com/OpenHelix-Team/Spatial-Forcing/tree/master/SLA}{\githubicon\,\textcolor[HTML]{1F4E8C}{GitHub}}}
\metadata[Project Page]{\href{https://sla-team.github.io/omnisf_projectpage/}{\projecticon\,\textcolor[HTML]{1F4E8C}{Project Website}}}

\begin{document}
\maketitle

\begin{figure}[H]
  \centering
  \includegraphics[width=0.99\textwidth]{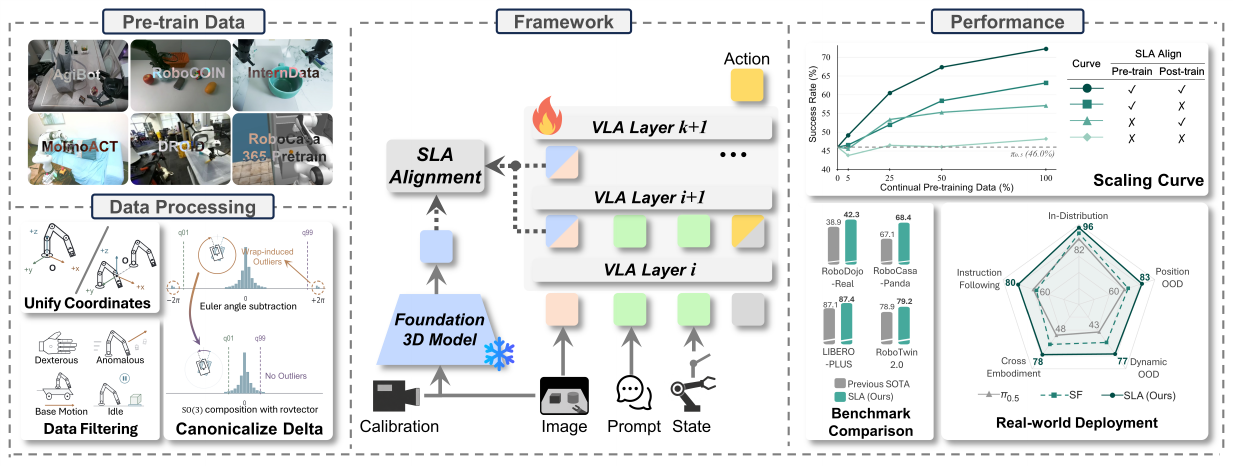}
  \caption{\method{} unifies heterogeneous robot data, aligns intermediate visual embeddings with frozen 3D features during training, and improves benchmark and real-world manipulation performance.}
  \label{fig:teaser}
\end{figure}

\clearpage
\setcounter{tocdepth}{3}
\tableofcontents
\clearpage

\section{Introduction}



The development of robotic manipulation hinges on integrating semantic reasoning with accurate physical control in the 3D real world.
Recently, vision-language-action (VLA) models~\citep{pi0,pi05,qwenrobotmanip,xiaomi_robotics1,g05,univla,song2026reconvla}, capable of instruction following and robotic action execution, have attracted significant attention.
By combining the pretrained vision-language model (VLM) with action experts learning on diverse robot demonstrations, these models offer a promising foundation for general-purpose manipulation.
However, successful manipulation requires not only recognizing objects and understanding instructions but also executing precise actions in the 3D physical world.
Typical 2D VLA foundation models inherit knowledge from VLMs pretrained on 2D image--text pairs and learn robot control without explicitly incorporating 3D physical knowledge during pre-training.
This limits their reliability in scenes that demand fine-grained spatial reasoning.

A direct approach to improving spatial understanding is to introduce 3D modalities during post-training on top of a pretrained 2D VLA foundation model.
Several approaches~\citep{pointvla,geovla,bridgevla} use depth maps or point clouds obtained from depth cameras or other 3D sensors, encode their geometric information, and feed it into the VLM backbone or action head.
However, some limitations make it difficult to develop a universal and scalable 3D strategy for this paradigm:
(1) Reliable 3D sensor data is hard to acquire, as measured depth maps and point clouds can be noisy or incomplete.
(2) Robotic sensors vary in types, positions, and calibration status, which introduces substantial heterogeneity into the data.
(3) Portions of existing large-scale robot datasets~\citep{oxe,robocoin,molmoact,bridgev2,fractal} do not include depth information, which limits the potential of scaling.

To address these questions, Spatial Forcing~\citep{li2025spatial} has propose a implicit 3D representation alignment strategy to enbow 2D VLA with spatial intelligence without explicit demand of depth information.
However, Spatial Forcing still concentrate on post-training, leaving spatial competence as a task-specific add-on and limiting the development of scalable 3D VLA foundation models with transferable spatial understanding.

Recently, continual pre-training has been widely adopted as a means of developing specific capabilities at scale while preserving the knowledge acquired during foundation model pre-training~\citep{groundingpi,vpp2}. Motivated by this approach, we propose to build on the strong 2D foundation model and develop spatial understanding through continual pre-training, laying the foundation for subsequent adaptation.
We therefore propose \textbf{\method{}}, 
a robotic foundation model with generalized spatial understanding capabilities developed through implicit representation alignment during both continual pre-training and post-training.
Intuitively, action generation is conditioned on visual and linguistic representations, making the spatial content of visual tokens directly relevant to action. Higher-quality visual tokens that contain richer spatial information can facilitate generating more precise manipulation actions.
Inspired by this, we align the intermediate visual embeddings of the VLA with external spatial representations extracted from a frozen 3D perception foundation model.
This spatial alignment objective complements the standard action-learning objective, encouraging the policy to internalize geometric information in the representations used for action prediction.
Crucially, this supervision requires no changes to the pretrained VLA architecture or the action-generation head.
We can therefore initialize the policy directly from the pretrained $\pi_{0.5}$ foundation~\citep{pi05} and continue training from its learned robotic knowledges.
This architectural compatibility allows spatial supervision to shape the foundation model while retaining access to the general knowledge encoded in its pretrained weights.

Technically, to support continual pre-training across heterogeneous robot data, we combines a unified training corpus.
We curate multiple large-scale public datasets covering diverse embodiments, tasks, scenes, and visual appearances, and unify their state and action annotations through consistent end-effector representations and coordinate conventions.
This processing provides a common basis for learning how different robots act in space.
The resulting visual diversity also makes a single manually chosen alignment layer restrictive: a certain intermediate visual layer of VLA models suited to one domain data may not be equally suitable for another.
We therefore develop adaptive layer selection mechanism, using feature-dependent gates to aggregate visual embeddings from multiple VLA layers for spatial supervision.
In addition, we adopt OmniVGGT~\citep{omnivggt} as the external 3D perception foundation to exploit partially available camera calibrations and depth annotations alongside RGB images.
We further introduce probabilistic state dropout during training to mitigate overfitting to demonstration trajectories and encourage the policy to make better use of visual cues, thereby improving generalization.
After continual pre-training, we retain spatial alignment during post-training to adapt the resulting foundation model to specific down-streaming tasks.
The teacher, alignment projector, and layer-selection gates are used only during training.
At deployment, the policy retains its original inference architecture without requiring the external perception model, alignment projector, or layer-selection gates.

Extensive evaluations across three public simulation benchmarks, one public real-world benchmark, and eight real-world deployment conditions are conducted to demonstrate the effectiveness of our \method{}.
On LIBERO-Plus~\citep{fei2025libero}, it achieves an overall success rate of 87.4\% under diverse observation, language, and scene perturbations, outperforming the compared baselines.
On RoboTwin 2.0~\citep{chen2025robotwin}, training on clean demonstrations yields a 72.3\% success rate in the Clean2Random setting and a 79.2\% average across Clean2Clean and Clean2Random, demonstrating generalization beyond the training conditions.
On RoboCasa-Panda~\citep{robocasa}, \method{} achieves 68.4\% success with only 50 downstream demonstrations per task, exceeding compared policies trained with 300--3,000 demonstrations per task.
On RoboDojo-Real~\citep{robodojo-leaderboard}, a single policy adapted across three embodiments achieves 33.9\% overall success, compared with 12.8\% for $\pi_{0.5}$, supporting its applicability to real robot manipulation.
Our eight deployment conditions further examine bimanual coordination, spatial and dynamic generalization, few-shot cross-embodiment adaptation, and instruction following.
\method{} achieves the highest success rate among the compared methods in every condition, averaging 83.8\% and improving over $\pi_{0.5}$ by 24.0 percentage points.
Controlled data-scaling experiments further show that applying spatial alignment in both training stages outperforms either single-stage alternative at every tested data proportion.
Together, these results support developing spatial understanding during continual pre-training and stengthen it during downstream adaptation.

Our contributions are summarized as follows:

\begin{itemize}
    \item We introduce \method{}, a robotic foundation model that implicitly enforces the integration of visual embeddings in VLAs with external spatial representations.
    \item We extend \method{} to continual pre-training on a carefully curated and unified corpus of public robot datasets, developing a spatially informed foundation model.
    \item We apply \method{} training principle during downstream post-training and demonstrate strong benchmark performance, together with bimanual coordination, spatial and dynamic generalization, few-shot cross-embodiment adaptation, and instruction following.
\end{itemize}

\section{Related Work}
\label{sec:relatedwork}

\subsection{VLA Foundation Models}
VLA foundation models learn generalist manipulation policies from large, diverse datasets and adapt them to downstream tasks. Cross-embodiment datasets~\citep{oxe,bridgev2,fractal,robomind,rh20t,abc130k,galaxeaopenworld} have accelerated this development. OpenVLA~\citep{openvla}, $\pi_0$~\citep{pi0}, RDT-1B~\citep{liu2024rdt}, and GR00T N1~\citep{bjorck2025gr00t} learn language-conditioned action prediction from robot demonstrations, while GR-2~\citep{gr2} combines large-scale video pre-training with policy learning on robot trajectories. To improve task understanding, RT-H~\citep{Rt-h} and $\pi_{0.5}$~\citep{pi05} adopt language-conditioned task hierarchies, whereas MolmoAct~\citep{molmoact} and MolmoAct2~\citep{molmoact2} incorporate action reasoning into robot control. RT-2~\citep{Rt-2}, $\pi_{0.5}$, and Xiaomi-Robotics-0~\citep{xiaomi_robotics0} also co-train on robot and vision-language data to support semantic generalization and instruction following. The scarcity of teleoperated robot data has further motivated alternative sources of supervision. HumanScale~\citep{humanscale} investigates pre-training on egocentric human videos followed by adaptation with robot data. Separately, the Universal Manipulation Interface (UMI)~\citep{umi} enables portable demonstration collection using handheld grippers without requiring robots at collection sites. RDT2~\citep{rdt2} and Xiaomi-Robotics-1~\citep{xiaomi_robotics1} exploit UMI data to expand training scale and coverage of environments and tasks. These advances support generalist action learning, but data scaling alone does not specify how geometric representations should be supervised during pre-training. Our complementary goal is to develop 3D spatial understanding in an existing VLA foundation model through spatially supervised continual pre-training.

\subsection{3D VLA Models}
Existing 3D VLA methods introduce geometry through explicit inputs, feature fusion, or representation supervision. PointVLA~\citep{pointvla} and GeoVLA~\citep{geovla} integrate point cloud features into action prediction. BridgeVLA~\citep{bridgevla} projects 3D observations into multi-view images; and FALCON~\citep{zhang2025falcon} extracts geometric tokens from spatial foundation models for fusion within the action head. These approaches still require additional geometric input extraction at deployment. 
Another works use training-time supervision to encode spatial structure in internal policy representations. 
ROCKET~\citep{rocket} explores multilayer spatial alignment, VEGA~\citep{vega} supervises visual encoder outputs, and Mind-VLA~\citep{mindvla} introduces instruction-aware alignment of target-object representations. Such approaches emphasize adding 3D capabilities during post-training on top of generic 2D VLA pre-training~\citep{pi05,univla,openvla-oft}, leaving spatial understanding as a supplementary capability rather than establishing a scalable 3D-aware foundation. Recent studies incorporate geometry into pre-training through egocentric spatial embeddings~\citep{spatialvla}, egocentric 3D annotations~\citep{vipa-vla}, and object pose estimation~\citep{posevla}. However, these 3D-aware pre-training efforts remain relatively limited in scale compared with large-scale 2D VLA pre-training. In contrast, we build on pretrained 2D VLA through continual pre-training to develop generalizable 3D-aware foundation models.


\subsection{Continual Pre-training}
Continual pre-training extends the domain capabilities of existing foundation models through further large-scale pre-training, providing a basis for downstream adaptation. Early studies~\citep{BioBERT, ClinicalBERT, FinBERT, LEGAL-BERT, Dont-Stop-Pretraining} focus on adapting to domain-specific data distributions. Llemma~\citep{llemma} strengthens mathematical reasoning through continual pre-training on mathematical text and code. AgentFounder~\citep{agentfounder} uses diverse reasoning and action data to establish a foundation for agent capabilities before post-training. Recent efforts extend this paradigm to robotics: ET-VLA~\citep{et-vla} uses synthetic data for continual pre-training to transfer existing VLAs to new embodiments, whereas AXIS~\citep{axis} investigates how policy performance scales with data. These VLA efforts primarily focus on training data sources. In contrast, we use implicit spatial alignment framework during continual pre-training to develop generalizable 3D spatial understanding while preserving the pretrained VLA backbone architecture.

\section{Methodology}
\method{} aims to develop a scalable 3D VLA foundation model through continual pre-training. Starting from a pretrained 2D VLA backbone~\citep{pi05}, we use external spatial representation supervision to shape its visual representations while retaining its backbone and action-generation architecture. \Cref{sec:motivation} motivates this approach, \Cref{sec:model} presents the model and training mechanisms, and \Cref{sec:data} describes the continual-pre-training corpus and its unified processing.

\subsection{Motivation}
\label{sec:motivation}
VLA models are built upon pretrained VLMs and generate actions through specialized designs for action experts. A $\theta$-parameterized VLM model employs multiple causal attention layers in an auto-regressive manner to generate the next tokens, formed as $\bm{x}_t \sim p_\theta(\bm{x}_t \mid \bm{x}_{<t}),$ where $\{\bm{x}_t\}_{t=1}^T$ denotes a sequence of tokens.

When adapted to VLA models, three modalities of information will be processed: vision, text, and action. The vision modality consists of multi-view images captured by robots, which are transformed into $N$ visual tokens $\{\bm{x}_t^\mathcal{V}\}_{t=1}^N$ through pretrained visual encoders such as DINOv2 ~\citep{dinov2} or SigLIP~\citep{siglip}. The text modality consists of task instructions, which are converted into $M$ linguistic tokens $\{\bm{x}_t^\mathcal{L}\}_{t=1}^M$ by a text tokenizer. Then, the VLA models generate $K$ action tokens $\{\bm{x}_t^\mathcal{A}\}_{t=1}^K$ conditioned on the preceding visual and linguistic tokens:
\begin{equation}
\label{eq:action}
\bm{x}_t^\mathcal{A} \sim p_\theta \big(\bm{x}_t^\mathcal{A} \mid \{\bm{x}_i^\mathcal{V}\}_{i=1}^N, \{\bm{x}_j^\mathcal{L}\}_{j=1}^M, \bm{x}_{<t}^\mathcal{A} \big),
\end{equation}
\begin{equation}
\label{eq:action_loss}
    \begin{aligned}
    \mathcal{L}_{\mathrm{action}} = \mathcal{L}[\mathcal{G}(\{\bm{x}_t^\mathcal{A}\}_{t=1}^K), A_{gt}],
    \end{aligned}
\end{equation}
where $\mathcal{L}[\cdot,\cdot]$ denotes the training loss (\textit{e.g.} L1, L2, or cross-entropy loss), $\mathcal{G}$ denotes the trainable action expert (\textit{e.g.} multi-layer MLP or flow-matching head~\citep{flow-matching}). 
\Cref{eq:action} illustrates that the visual tokens as intermediate scene representations play a crucial role in generating action tokens and should be supervised properly.

Since VLA models must capture 3D information and reason about the relative positions between robots and objects to generate precise spatial movements, we hypothesize that the 3D information is implicitly embedded within the visual embeddings of VLAs. Such embeddings would allow action tokens to acquire 3D cues through the auto-regressive mechanism during inference.

\begin{figure*}[t]
    \centering
    \includegraphics[pagebox=cropbox,width=0.98\textwidth]{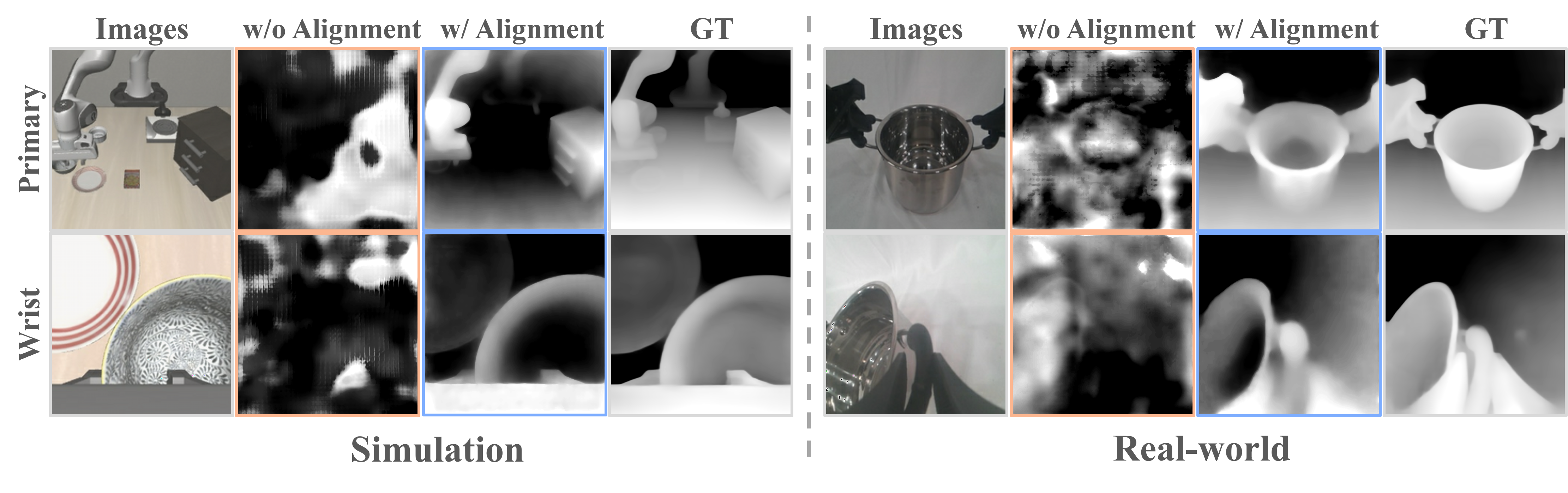}
    \caption{\textbf{Depth probing of the visual embeddings of VLAs.} Embeddings learned solely from 2D images without alignment do not produce meaningful spatial structures. The aligned embeddings inherently contain rich spatial information, leading to better performance in depth probing.}
    \label{fig:depth_probing}
\end{figure*}

To evaluate this hypothesis, we follow linear probing~\citep{linearprobing,geometry-forcing} to conduct a lightweight depth probing experiment. Specifically, we freeze all the parameters of a VLA model~\citep{openvla-oft} and only train a DPT head~\citep{dpt} to transform the visual embeddings of VLA to depth maps. This enables us to quantify the richness of spatial information embedded in the VLA representation space. As shown in \Cref{fig:depth_probing}, the probing results show that visual embeddings learned solely from 2D images do not produce meaningful spatial structures, suggesting a limited capacity of the model to encode 3D structure information without explicit spatial input or guidance.

However, bridging this 3D physical gap through additional sensor inputs or depth-estimation modules would change the architecture of the original 2D VLA foundation model and may disrupt the general knowledge already acquired during its pre-training. We therefore seek a training strategy that reuses the existing VLA architecture and pretrained parameters to unlock its generalizable understanding of the 3D physical world through continual pre-training.

\subsection{Model}
\label{sec:model}

\subsubsection{Model Architecture}

Our main philosophy is to employ external supervision signals to supervise the visual tokens $\bm{x}^{\mathcal V}$. To provide the signals with rich spatial information, we first input a set of multi-view images $\mathcal I$ into a frozen external 3D perception foundation model $f^{\mathrm{3D}}$, which outputs pixel-level 3D representations $f^{\mathrm{3D}}(\mathcal I)$ for each image. Available geometric annotations like camera parameters and depth images are also provided to this external 3D foundation model, as described in \Cref{sec:external_spatial}. Additionally, these spatial representations are augmented with positional embeddings $E$ to ensure that the supervised tokens preserve the critical spatial position order during supervised training. We extract the final-layer representations of $f^{\mathrm{3D}}(\mathcal I)$ and bilinearly resample the feature grids with positional embeddings to the VLA token resolution, obtaining a spatial target $\mathbf y_i$ at each visual token location:
\begin{equation}
    \mathbf y=\mathcal R\!\left[f^{\mathrm{3D}}(\mathcal I)+E\right],
    \label{eq:spatial_target}
\end{equation}
where $\mathcal R\left[\cdot\right]$ denotes bilinear resampling to the VLA visual token resolution. For brevity, the available geometric annotations like camera parameters are implicit in $\mathcal I$.

To align the VLA's per-pixel visual tokens with this external 3D perception foundation model, we first process each VLA visual token $\bm{x}^{\mathcal V}$ with layer normalization $\mathrm{LN}$ followed by a two-layer MLP to ensure compatibility in feature dimension. Let ${}^{(\ell)}\bm{x}^{\mathcal V}$ denote the VLA visual tokens after attention layer $\ell$ of the VLA's vision-language backbone. The normalization and MLP parameters are shared across layers. As there are multiple causal attention layers in the VLA model, visual tokens can be obtained after different layers. We adaptively aggregate their projected features into $\mathbf z$:
\begin{equation}
    \begin{aligned}
        \mathbf z
        &=\mathcal F\!\left[\left\{{}^{(\ell)}\bm{x}^{\mathcal V}\right\}_{\ell=1}^{L}\right]\\
        &=\sum_{\ell=1}^{L}{}^{(\ell)}\mathbf g\odot
        \mathrm{MLP}\!\left[\mathrm{LN}\!\left({}^{(\ell)}\bm{x}^{\mathcal V}\right)\right],
    \end{aligned}
    \label{eq:layer_fusion}
\end{equation}
where $L$ is the number of VLA attention layers, ${}^{(\ell)}\mathbf g_i$ is a token- and channel-wise gate, and $\odot$ denotes element-wise multiplication. This fusion operator $\mathcal F\left[\cdot\right]$ denotes the adaptive layer selection strategy, which is detailed in \Cref{sec:adaptive_layers}.

Then we employ a cosine similarity score to maximize the alignment between the visual tokens of the VLA and the external spatial representation signals. The alignment loss is
\begin{equation}
    \mathcal L_{\mathrm{align}}
    =\frac{1}{N}\sum_{i=1}^{N}\left[1-\mathcal S(\mathbf z_i,\mathbf y_i)\right],
    \label{eq:align_loss}
\end{equation}
where $\mathcal S\left[\cdot,\cdot\right]$ denotes cosine similarity and $N$ is the number of visual tokens across views. The target $\mathbf y_i$ corresponds to the same view and image location as visual token $\mathbf z_i$. Gradients update the VLA, MLP projector, and aggregation gates, while the external 3D perception model $f^{\mathrm{3D}}$ remains frozen.

The final training objective combines both the standard training loss for action generation and the 3D foundation model alignment loss with a weighting factor $\alpha$:
\begin{equation}
    \mathcal L=\mathcal L_{\mathrm{action}}+\alpha\mathcal L_{\mathrm{align}},
    \label{eq:total_loss}
\end{equation}
where $\alpha$ controls the spatial alignment strength. We instantiate \method{} with the pretrained $\pi_{0.5}$ policy~\citep{pi05}, whose action expert generates continuous action chunks through flow matching.

During inference, the VLA model trained in the \method{} manner operates identically to the original VLA, introducing no additional structures or computational overhead. The frozen 3D foundation model, alignment projector, and adaptive layer gates are used only during training and are discarded at inference. Neither depth maps nor camera calibration are required as additional policy inputs for inference.

\subsubsection{External Spatial Supervision}
\label{sec:external_spatial}

The external 3D perception foundation model $f^{\mathrm{3D}}$ provides the spatial representations as alignment signals. Apart from RGB image observations, some source datasets provide calibrated camera parameters and depth images, while others provide only partial geometric annotations or none at all. Discarding these annotations would leave useful geometric constraints unused, whereas requiring complete annotations would exclude otherwise valuable demonstrations. Therefore, the external model should accommodate the uneven availability of geometric annotations across the continual-pre-training corpus.

We adopt OmniVGGT~\citep{omnivggt} as the external spatial model. Built on VGGT~\citep{vggt}, it alternates frame-wise and cross-view attention to integrate local structure with relationships across images. Its GeoAdapter encodes available camera intrinsics, extrinsics, and depth maps through separate camera and depth branches, using placeholder tokens when annotations are missing. Stochastic multimodal fusion exposes the model to different combinations of geometric inputs, including RGB-only observations. These mechanisms combine visual evidence and arbitrary available geometric constraints in a unified latent representation. We therefore argue that its backbone features contain sufficiently rich spatial information to serve as external supervision for the VLA's visual tokens.

\subsubsection{Adaptive Layer Selection}
\label{sec:adaptive_layers}

A heterogeneous pre-training corpus makes a fixed alignment layer a restrictive choice. Manually selecting a single layer requires separate training experiments, and a layer selected for one data domain need not offer the same balance of visual detail and task context in another. Differences in scenes, textures, and embodiments motivate a mechanism that can adapt the contribution of each layer to the observed visual content.

Inspired by Gated Attention~\citep{gated-attention}, we learn a feature-dependent gate for every visual token at every VLA transformer layer. Given the unprojected embedding ${}^{(\ell)}\bm{x}_i^{\mathcal V}$, the gate is
\begin{equation}
    {}^{(\ell)}\mathbf g_i
        =\sigma\!\left(W_g\,{}^{(\ell)}\bm{x}_i^{\mathcal V}+\mathbf b_g\right),
    \label{eq:adaptive_layers}
\end{equation}
where $W_g$ and $\mathbf b_g$ are the learnable weight matrix and bias vector of the gate, respectively, and $\sigma$ is the element-wise sigmoid. These gate parameters are shared across layers. The fusion operator $\mathcal F$ in \Cref{eq:layer_fusion} uses these gates to weight the projected features and sums them across layers to obtain $\mathbf z_i$.

Through this mechanism, multi-layer VLA visual tokens can be supervised simultaneously, allowing different visual cues like depth and channel informations contribute to a unified spatial representation. The gates belong exclusively to the auxiliary training branch and introduce no additional policy computation at deployment.

\subsubsection{State Dropout}
\label{sec:state_dropout}

Spatial supervision is useful only when the policy learns to rely on visual evidence for control. Robot states provide precise proprioceptive information, but can become shortcuts: a policy may associate particular states with demonstrated trajectories instead of responding to visual changes in the scene~\citep{GAP,state-free-policy}. We apply state dropout to discourage this shortcut during both training stages. Independently for each training sample, we randomly omit the entire state-token segment during tokenization; the instruction and image tokens are retained. Together, our external spatial alignment and state dropout encourage the model to encode scene geometry and to use visual observations when generating actions. This dropout is disabled at inference when the normal state input is available.

\subsection{Data}
\label{sec:data}

The quality and diversity of training data are fundamental to the generalization ability of VLAs. To build a VLA model that generalizes well across embodiments, tasks, and environments, we carefully curate a large-scale corpus for continual pre-training. Furthermore, we apply a unified data curation and preprocessing pipeline to all sources to ensure consistent robot state and action annotations.

\subsubsection{Data Sources}

The six source datasets summarized in \Cref{tab:pretraining_data} collectively contain approximately 1.7 million episodes and 14,330 hours of demonstrations. 

\begin{table}[t]
    \centering
    \caption{Overview of the source datasets for continual pre-training. Statistics refer to the source datasets before our data selection and processing; task counts are summed across datasets.}
    \label{tab:pretraining_data}
    \begin{tabular}{cccc}
        \toprule
        Data source & Episodes & Tasks & Hours\\
        \midrule
        InternData-A1~\citep{interndata-a1} & 630K & 70 & 7430 \\
        AgiBotWorld-Beta~\citep{agibot-world} & 168K & 217 & 2970 \\
        RoboCasa365-Pretrain~\citep{robocasa365} & 630K & 300 & 2090 \\
        RoboCOIN~\citep{robocoin} & 183K & 808 & 1300 \\
        DROID~\citep{droid} & 95K & 86 & 510 \\
        MolmoAct~\citep{molmoact} & 7.9K & 23 & 30 \\
        \midrule
        \textbf{Total} & \textbf{1,713.9K} & \textbf{1,504} & \textbf{14,330} \\
        \bottomrule
    \end{tabular}
\end{table}

\textit{RoboCOIN}~\citep{robocoin} provides real-world bimanual manipulation demonstrations collected across heterogeneous robotic platforms, adding variation in embodiment and observation configuration.

\textit{AgiBot World Beta}~\citep{agibot-world} offers a large collection of real-world manipulation demonstrations with broad coverage of tasks and scenes. We retain its two-finger-gripper demonstrations without mobile-base motion.

\textit{DROID}~\citep{droid} supplies demonstrations collected with Franka robots in varied real-world environments. Demonstration quality and annotation precision vary somewhat. We include it for its single-arm real-robot experience, which remains comparatively scarce in public datasets.

\textit{InternData-A1}~\citep{interndata-a1} provides synthetic manipulation trajectories generated by scripted policies, which tend to produce regular, recurring motion patterns. We use its Split ALOHA, Lift2, and Franka subsets for their rich variation in simulated scenes, assets, and textures, broadening visual coverage.

\textit{MolmoAct}~\citep{molmoact} contributes real-world franka demonstrations centered on a focused set of tasks and scenes. It further complements valuable single-arm real-robot experience to the corpus.

\textit{RoboCasa365}~\citep{robocasa365} provides demonstrations of household tasks in simulated kitchens. We use its pre-training atomic, composite, and MimicGen subsets after removing trajectories with base motion. Its post-train subsets are not used.

To ensure fair and valid comparisons after post-training, our continual pre-training corpus excludes downstream benchmark data that would cause evaluation leakage.

\subsubsection{Data Processing}
\label{sec:data_processing}

\paragraph{State and Action Representation}

Different robot morphologies induce incompatible joint-state and joint-action spaces, making joint coordinates a difficult basis for shared control representations. We instead express action through end-effector (EEF) poses, which describe the spatial effect of an action independently of the number and arrangement of arm joints. We use absolute EEF states as observations and chunk-wise delta EEF actions as prediction targets.

We first express all poses in a consistent robot-centered EEF coordinate frame. We unified the coordinate origin to the arm base for single-arm robots and the midpoint between the two arm bases for bimanual robots. Besides, we unified its axes point to forward ($+x$), left ($+y$), and upward ($+z$). A rigid coordinate transformation maps each source data's EEF positions and orientations into this convention. Thus, the numerical axes of the shared action representation have consistent physical meanings across embodiments.

\begin{figure}[t]
    \centering
    \includegraphics[width=0.75\columnwidth]{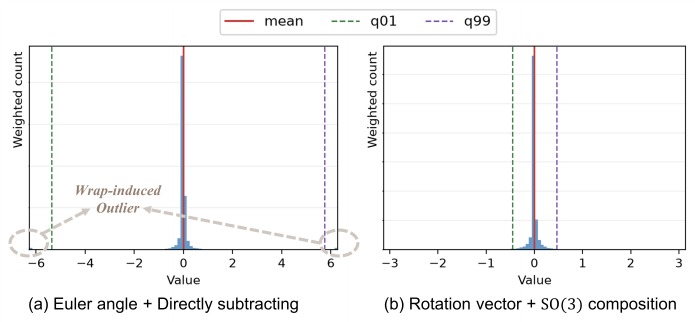}
    \caption{\textbf{Comparison of rotational action distributions.} \textbf{(a)} Euler-angle representations with direct subtraction for relative-rotation targets. \textbf{(b)} Rotation-vector representations with relative rotations computed through $\mathrm{SO}(3)$ composition. In (a), angular wrapping produces outliers near $\pm2\pi$, which widen the interval between the 1st and 99th percentiles. This severely compresses the range of typical action values after normalization, hindering training.}
    \label{fig:rotation_distribution}
\end{figure}

Rotation format is equally important for consistent action targets. Common choices include Euler angles, continuous 6D representations, rotation vectors, and quaternions. Directly subtracting Euler angles as delta actions may convert a small physical rotation across an angular wrap boundary into a large numerical displacement. We observe such wrap-induced outliers in the pre-training action distributions shown in \Cref{fig:rotation_distribution}. Our controlled comparisons motivate representing orientation with rotation vectors while computing relative rotations through $\mathrm{SO}(3)$ composition.

Specifically, for one arm, let $(\mathbf p_0,R_0)$ be its current absolute EEF pose and $(\mathbf p_t,R_t)$ the absolute target pose of action $t\in\{1,\ldots,H\}$ within a chunk of $H$ actions. We construct
\begin{align}
    \Delta\mathbf p_t&=\mathbf p_t-\mathbf p_0,\\
    \Delta R_t&=R_tR_0^\top,\\
    \Delta\mathbf r_t&=\operatorname{Log}(\Delta R_t)^\vee,
    \label{eq:relative_action}
\end{align}
where $\operatorname{Log}(\cdot)^\vee$ maps a relative rotation matrix to its rotation-vector representation. Every action in the chunk uses the same current pose as its reference, rather than the preceding action.

A rotation vector has a unified physical meaning through its direction and magnitude, making it natural to normalize its three channels using a shared scale. However, our controlled experiments show that channel-wise normalization yields substantially better performance, and we therefore adopt this approach in our experiments.

\paragraph{Downsampling}

Large numbers of similar trajectories can dominate a mixed training corpus without adding comparable behavioral diversity. This is particularly relevant to scripted simulation data, where repeated executions may follow highly regular motion patterns. We reduce such redundancy by subsampling complete episodes within tasks, preserving the original temporal sampling rate of every retained trajectory.

For example, for InternData-A1, we retain 3\% of episodes for selected tasks with highly repetitive trajectories, and 10\% for the other tasks. For the pre-training subsets of RoboCasa365, we retain 0.5\% of the MimicGen episodes and 15\% of the composite episodes. These source-dependent and task-dependent retention rates bring episode counts closer to a comparable order of magnitude, reducing the dominance of a small number of scenes or tasks.

\paragraph{Filtering}

We focus on manipulation with two-finger grippers and exclude dexterous-hand data and trajectories involving robot-base motion. We additionally use action statistics to identify and remove data with anomalous arm-motion speeds. Where existing annotations are available~\citep{ye2026world}, we remove idle frames in which both the robot and the scene remain stationary. Together, these filters keep the corpus aligned with the intended control setting and reduce uninformative or inconsistent supervision.

\section{Experiments}
\label{sec:exp}

\subsection{Experimental Setup}
\label{sec:exp_setup}


We evaluate \method{} through benchmark-specific post-training on three simulation benchmarks, one real-world benchmark, and plenty real-world deployments. These evaluations cover robustness to visual and language perturbations, manipulation under scene variation, and applicability across robot embodiments.

LIBERO-Plus~\citep{fei2025libero} tests robustness along seven perturbation dimensions: camera, robot, language, lighting, background, observation noise, and object layout. We post-train on the standard LIBERO demonstrations and evaluate the perturbed environments without additional training on LIBERO-Plus. Evaluation comprises 10,030 rollouts; the overall SR weights each dimension by its number of evaluation episodes.

RoboTwin 2.0~\citep{chen2025robotwin} contains 50 bimanual manipulation tasks. We jointly post-train on 50 clean demonstrations per task and evaluate each task for 100 trials in both the Clean2Clean and Clean2Random settings. Clean2Random introduces domain randomization to test generalization beyond the clean training conditions. We report the task-averaged SR for each setting and their equal-weight average.

RoboCasa-Panda~\citep{robocasa} contains 24-task kitchen manipulation benchmark covering pick-and-place, articulated-object interaction, and appliance operation with a Panda robot. Our post-training uses 50 demonstrations per task. Each task is evaluated for 50 trials under each of three evaluation seeds, yielding 3,600 trials in total. Results are averaged over seeds and tasks.

RoboDojo-Real~\citep{robodojo-leaderboard} covers 18 tasks across ARX X5, Piper, and Piper X, with six tasks per embodiment. We jointly post-train a single model on demonstrations from all three embodiments. We report scores on a 0--100 scale to measure partial task progress and success rates to measure complete task success. Aggregation equally weights tasks within each embodiment and then the three embodiments.


Apart from these benchmark evaluation, we additionally conduct real-world deployment experiments under eight private deployment conditions A--H on a bimanual Piper platform and a single-arm Franka robot. Piper platform uses one main camera and two wrist cameras while Franka uses one main camera and one wrist camera. The type of All cameras is Realsense D435. For each method, we post-train one policy on 100 demonstrations of A and reuse it for A--C evaluation. For D, we jointly post-train on the 100 Piper demonstrations of A and 10 additional Franka demonstrations. For task E--H, one policy is post-trained on 100 demonstrations collected in the E scene configuration and evaluated across E--H. We report object-level successs rates over 50 evaluated object instances per method and condition. For A, B, and D, these instances come from 10 rollouts with five objects per rollout, and each object is scored separately. C and E--H each comprise 50 single-object rollouts. Success requires complete manipulation sequence for the evaluated object.

\subsection{Implementation Details}

We initialize \method{} from the pretrained $\pi_{0.5}$ policy~\citep{pi05} to conduct continual pre-training, followed by post-training for specific tasks. 

The continual pre-training runs for 36k optimizer steps with an effective global batch size of 4,608, obtained by accumulating gradients over two microsteps with a global batch size of 2,304 each. The learning rate warms up to $5\times10^{-5}$ over the first 1k steps, follows cosine decay to $1\times10^{-5}$ at step 30k, and remains constant thereafter. Continual pre-training takes approximately 110 hours on 48 NVIDIA H100 GPUs.

As for post-training, LIBERO-Plus takes 512 batch size and 30k steps, RoboTwin takes 256 batch size and 30k steps, RoboCasa-Panda takes 64 batch size and 60k steps, RoboDojo-Real takes 512 batch size and 40k steps, and all real-world deployment experiments take 64 batch size and 20k steps.

\subsection{Comparisons with State-of-The-Art Methods}
We compare against the methods listed in the benchmark tables. Spatial Forcing denotes our original method applied only during post-training from pretrained $\pi_{0.5}$, without our continual pre-training stage.

\subsubsection{LIBERO-Plus}
\begin{table*}[!htp]
    \centering
    \caption{\textbf{LIBERO-Plus benchmark success rates (\%, $\uparrow$).} Models are fine-tuned on the standard LIBERO training set without LIBERO-Plus set. Average is weighted by the number of evaluation episodes. \textbf{Bold} and \underline{underline} denote the best and second-best displayed values, respectively.}
    \label{tab:libero_plus}
    \resizebox{\textwidth}{!}{%
    \begin{tabular}{lcccccccc}
\toprule
Method & Camera & Robot & Language & Light & Background & Noise & Layout & Average \\
\midrule
$\pi_0$~\citep{pi0} & 13.8 & 6.0 & 58.8 & 85.0 & 81.4 & 79.0 & 68.9 & 53.6 \\
$\pi_{0.5}$~\citep{pi05} & 78.4 & 73.6 & 80.8 & 96.2 & 94.1 & 89.0 & 84.5 & 84.4 \\
StarVLA~\citep{starvla} & 52.5 & 49.8 & 88.5 & 95.7 & 95.7 & 73.0 & 76.9 & 74.1 \\
4D-WAM~\citep{4d-wam} & 45.1 & 64.3 & \textbf{90.6} & 94.3 & 57.7 & 69.1 & 79.2 & 71.0 \\
Abot-M0~\citep{abot-m0} & 60.4 & 67.9 & 86.4 & 96.2 & 91.6 & 86.4 & 82.6 & 80.5 \\
Cosmos Policy~\citep{kim2026cosmos} & 75.8 & 63.3 & 81.7 & 96.5 & 88.9 & 92.7 & 82.2 & 82.2 \\
Being-H0.7~\citep{luo2026being} & \underline{82.0} & 59.0 & 82.8 & \underline{97.8} & 90.0 & \underline{93.5} & \textbf{88.5} & 84.8 \\
Qwen-RobotManip (robot-only)~\citep{yuan2026qwen} & 72.8 & \underline{78.2} & \underline{88.7} & 97.5 & \underline{97.6} & \textbf{95.4} & 85.4 & \underline{87.1} \\
\midrule
\rowcolor{black!6} Spatial Forcing & 70.2 & \textbf{79.3} & 85.6 & 96.8 & 96.2 & 86.0 & \underline{88.3} & 85.1 \\
\rowcolor{black!6} \method{} (Ours) & \textbf{82.2} & 73.8 & 83.5 & \textbf{98.8} & \textbf{98.0} & \textbf{95.4} & 86.2 & \textbf{87.4} \\
\bottomrule
    \end{tabular}}%
\end{table*}

\Cref{tab:libero_plus} shows that \method{} achieves a weighted SR of 87.4\%, improving over $\pi_{0.5}$ and Spatial Forcing by 3.0 and 2.3 percentage points, respectively. It exceeds the robot-only Qwen-RobotManip model~\citep{yuan2026qwen} by 0.3 percentage points. Relative to the baseline $\pi_{0.5}$, improvements occur in all seven perturbation dimensions, including observation noise (89.0\% to 95.4\%) and background changes (94.1\% to 98.0\%). \method{} also achieves the highest lighting-perturbation SR among the compared methods, at 98.8\%. These results indicate that the benefits extend across a range of perturbations.
These gains may stem in part from spatially informed visual representations that help the policy preserve task-relevant scene structure under challenging observation perturbations.

\subsubsection{RoboTwin}
\begin{table}[!htbp]
    \centering
    \caption{\textbf{RoboTwin 2.0 benchmark success rates (\%, $\uparrow$).} Models are fine-tuned jointly on the 50 standard tasks using 50 demonstrations from the Clean setting per task. Per-task results are shown in \Cref{tab:robotwin_per_task}. \textbf{Bold} and \underline{underline} denote the best and second-best displayed values, with ties treated equally.}
    \label{tab:robotwin}
    \setlength{\tabcolsep}{5pt}
    \begin{tabular}{lccc}
\toprule
Method & Clean2Clean & Clean2Random & Average \\
\midrule
Discrete Forcing~\citep{discreteforcing} & 59.3 & 11.2 & 35.3 \\
X-WAM~\citep{x-wam} & 70.0 & 25.8 & 47.9 \\
$\pi_{0.5}$~\citep{pi05} & 70.7 & 46.0 & 58.3 \\
4D-WAM~\citep{4d-wam} & 81.5 & 41.8 & 61.7 \\
WorldScape Policy 2.0~\citep{worldscape2} & 84.9 & 38.5 & 61.7 \\
GigaBrain-0.7~\citep{gigabrain07} & 66.8 & 67.9 & 67.4 \\
UniWAM~\citep{uniwam} & 75.1 & 67.6 & 71.3 \\
ViGAR~\citep{vigar} & 82.0 & 67.0 & 74.5 \\
ME-Dex-1.0~\citep{medex} & \textbf{89.6} & \underline{68.1} & \underline{78.9} \\
\midrule
\rowcolor{black!6} Spatial Forcing & 74.7 & 62.6 & 68.7 \\
\rowcolor{black!6} \method{} (Ours) & \underline{86.0} & \textbf{72.3} & \textbf{79.2} \\
\bottomrule
    \end{tabular}
\end{table}

\Cref{tab:robotwin} shows that \method{} achieves 86.0\% Clean2Clean SR, 72.3\% Clean2Random SR, and a 79.2\% average on RoboTwin 2.0. The Clean2Random and average results exceed $\pi_{0.5}$ by 26.3 and 20.9 percentage points, respectively.
Clean2Clean primarily assesses performance under conditions similar to the training demonstrations, whereas Clean2Random tests generalization under domain randomization and thus provides a more demanding measure of robustness to scene variation.
Compared with ME-Dex-1.0~\citep{medex}, the strongest competing method by average SR in the table, \method{} improves Clean2Random SR by 4.2 points while trailing its Clean2Clean SR.
The average also exceeds the original Spatial Forcing method by 10.5 points, supporting the effectiveness of the complete training approach beyond post-training-only spatial alignment.
The substantial gain in the Clean2Random setting suggests that \method{} helps the policy generalize beyond the clean training conditions to more varied manipulation scenes.

\subsubsection{RoboCasa-Panda}
\begin{table}[!htbp]
    \centering
    \caption{\textbf{RoboCasa-Panda benchmark success rates (\%, $\uparrow$).} Per-task results are shown in \Cref{tab:robocasa_per_task}. \textbf{Bold} and \underline{underline} denote the best and second-best displayed values.}
    \label{tab:robocasa_panda}
    \begin{tabular}{lcc}
\toprule
Method & \makecell{Training Demos\\per Task} & Average \\
\midrule
GR00T-N1~\citep{bjorck2025gr00t} & 300 & 49.6 \\
UVA~\citep{li2025UVA} & 50 & 50.0 \\
DP-VLA~\citep{dp-vla} & 3000 & 57.3 \\
DreamGen~\citep{jang2025dreamgen} & 300 & 57.6 \\
DUST~\citep{dust} & 300 & 58.5 \\
UWM~\citep{uwm} & 1000 & 60.8 \\
$\pi_0$~\citep{pi0} & 300 & 62.5 \\
GR00T-N1.5~\citep{gr00t-n15} & 300 & 64.1 \\
Video Policy~\citep{video-policy} & 300 & 66.0 \\
FLARE~\citep{flare} & 300 & 66.4 \\
HAMLET~\citep{hamlet} & 300 & 66.4 \\
Cosmos Policy~\citep{kim2026cosmos} & 50 & 67.1 \\
\midrule
\rowcolor{black!6} Spatial Forcing & 50 & \underline{67.6} \\
\rowcolor{black!6} \method{} (Ours) & 50 & \textbf{68.4} \\
\bottomrule
    \end{tabular}
\end{table}

\Cref{tab:robocasa_panda} shows that \method{} reaches a mean success rate of 68.4\% on RoboCasa-Panda, the highest result among the compared methods. It exceeds Cosmos Policy~\citep{kim2026cosmos} by 1.3 percentage points and Spatial Forcing by 0.8 points, with all three using 50 post-training demonstrations per task. Other models use different demonstration budgets, as explicitly reported in the table. This comparison shows that our proposed approach remains competitive when adaptation data are limited. The task-specific breakdown in \Cref{tab:robocasa_per_task} shows that \method{} achieves success rates above 80\% on 10 of the 24 tasks, including 100.0\% on TurnOffMicrowave.


\subsubsection{RoboDojo-Real}
\label{sec:dojo-real}

\begin{table*}[!tp]
    \centering
    \caption{\textbf{RoboDojo real-world benchmark results across three embodiments.} Score ($\uparrow$) captures partial task progress, and SR (\%, $\uparrow$) measures binary task completion. Per-task results are shown in \Cref{tab:robodojo_per_task}. \textbf{Bold} and \underline{underline} denote the best and second-best displayed values.}
    \label{tab:robodojo_real}
    \begin{tabular}{lcccccccc}
\toprule
\multirow{2}{*}{Method} & \multicolumn{2}{c}{ARX X5} & \multicolumn{2}{c}{Piper} & \multicolumn{2}{c}{Piper X} & \multicolumn{2}{c}{Average} \\
\cmidrule(lr){2-3}\cmidrule(lr){4-5}\cmidrule(lr){6-7}\cmidrule(l){8-9}
 & Score & SR & Score & SR & Score & SR & Score & SR \\
\midrule
Spirit v1.5~\citep{spirit-v15} & 0.0 & 0.0 & 4.0 & 1.7 & 0.7 & 0.0 & 1.6 & 0.6 \\
StarVLA~\citep{starvla} & 0.0 & 0.0 & 12.0 & 5.0 & 0.5 & 0.0 & 4.1 & 1.7 \\
$\pi_0$~\citep{pi0} & 8.1 & 0.0 & 9.0 & 5.0 & 0.3 & 0.0 & 5.8 & 1.7 \\
GR00T-N1.7~\citep{bjorck2025gr00t} & 9.3 & 1.7 & 6.8 & 3.3 & 1.4 & 0.0 & 5.9 & 1.7 \\
X-VLA~\citep{zheng2025x} & 6.8 & 1.7 & 15.9 & 8.3 & 0.1 & 0.0 & 7.6 & 3.3 \\
Xiaomi-Robotics-0~\citep{xiaomi_robotics0} & 15.0 & 8.3 & 7.6 & 3.3 & 1.2 & 0.0 & 7.9 & 3.9 \\
GalaxeaVLA~\citep{galaxeaopenworld} & 3.4 & 0.0 & 22.0 & 13.3 & 1.7 & 0.0 & 9.0 & 4.4 \\
InternVLA-A1~\citep{internvla-a1} & 10.0 & 3.3 & 23.2 & 18.3 & 2.7 & 0.0 & 12.0 & 7.2 \\
$\pi_{0.5}$~\citep{pi05} & 27.7 & 13.3 & 32.0 & 21.7 & 9.1 & 3.3 & 22.9 & 12.8 \\
OpenWAM-$\alpha$~\citep{openwam} & 39.0 & 23.3 & \underline{46.7} & \underline{36.7} & \textbf{27.2} & \textbf{13.3} & 37.6 & 24.4 \\
InternW0-$\Delta$~\citep{internw0} & \textbf{59.7} & \textbf{50.0} & 39.3 & 31.7 & \underline{17.8} & \textbf{13.3} & \underline{38.9} & \underline{31.7} \\
\midrule
\rowcolor{black!6} \method{} (Ours) & \underline{50.3} & \underline{36.7} & \textbf{60.3} & \textbf{55.0} & 16.3 & \underline{10.0} & \textbf{42.3} & \textbf{33.9} \\
\bottomrule
    \end{tabular}
\end{table*}

\Cref{tab:robodojo_real} evaluates whether our approach remains useful for real-world benchmarks, especially when a single policy is adapted jointly to several physical embodiments. 
To address long-horizon tasks and long-term memory requirements in RoboDojo-Real benchmark, we incorporate a memory mechanism into the policy during post-training, as detailed in \cref{sec:memory}.
\method{} achieves the highest overall score and success rate among the compared methods at 42.3 and 33.9\%, exceeding $\pi_{0.5}$ by 19.4 score points and 21.1 percentage points, respectively. It also surpasses InternW0-$\Delta$~\citep{internw0}, the strongest competing method overall, by 3.4 score points and 2.2 percentage points. On Piper, \method{} leads in both metrics, with a Score of 60.3 and SR of 55.0\%. On ARX X5, it ranks second in both metrics behind InternW0-$\Delta$.  These results support the effectiveness of our training approach to joint multi-embodiment adaptation,
highlighting its potential for real-world manipulation. The spatial alignment branch is used only during training, while the historical image inputs of the memory mechanism are retained at inference.


\subsection{Performance Analysis}

\subsubsection{Data Scaling Trend}
\label{sec:exp_scaling}

\begin{table}[!htbp]
    \centering
    \caption{\textbf{Continual pre-training data scaling ablation.} \method{} alignment is independently enabled or disabled during continual pre-training and post-training. The corresponding line chart is plotted in \Cref{fig:scaling}. Experiments are conducted under the RoboTwin Clean2Random settings to evaluate success rates (\%, $\uparrow$).}
    \label{tab:scaling}
    \setlength{\tabcolsep}{3pt}
    \begin{tabular}{*{6}{>{\centering\arraybackslash}m{0.145\linewidth}}}
\toprule
\multicolumn{2}{c}{\method{} Alignment} & \multicolumn{4}{c}{Continual Pre-training Data (\%)} \\
\cmidrule(lr){1-2}\cmidrule(l){3-6}
Pre-training & Post-training & 5 & 25 & 50 & 100 \\
\midrule
\Crossmark & \Crossmark & 43.8 & 46.5 & 46.1 & 48.2 \\
\Crossmark & \Checkmark & 45.6 & \underline{53.4} & 55.3 & 57.1 \\
\Checkmark & \Crossmark & \underline{46.6} & 52.0 & \underline{58.4} & \underline{63.2} \\
\Checkmark & \Checkmark & \textbf{49.2} & \textbf{60.5} & \textbf{67.4} & \textbf{72.3} \\
\bottomrule
    \end{tabular}
\end{table}

\begin{figure}[!htbp]
    \centering
    \includegraphics[width=0.6\columnwidth]{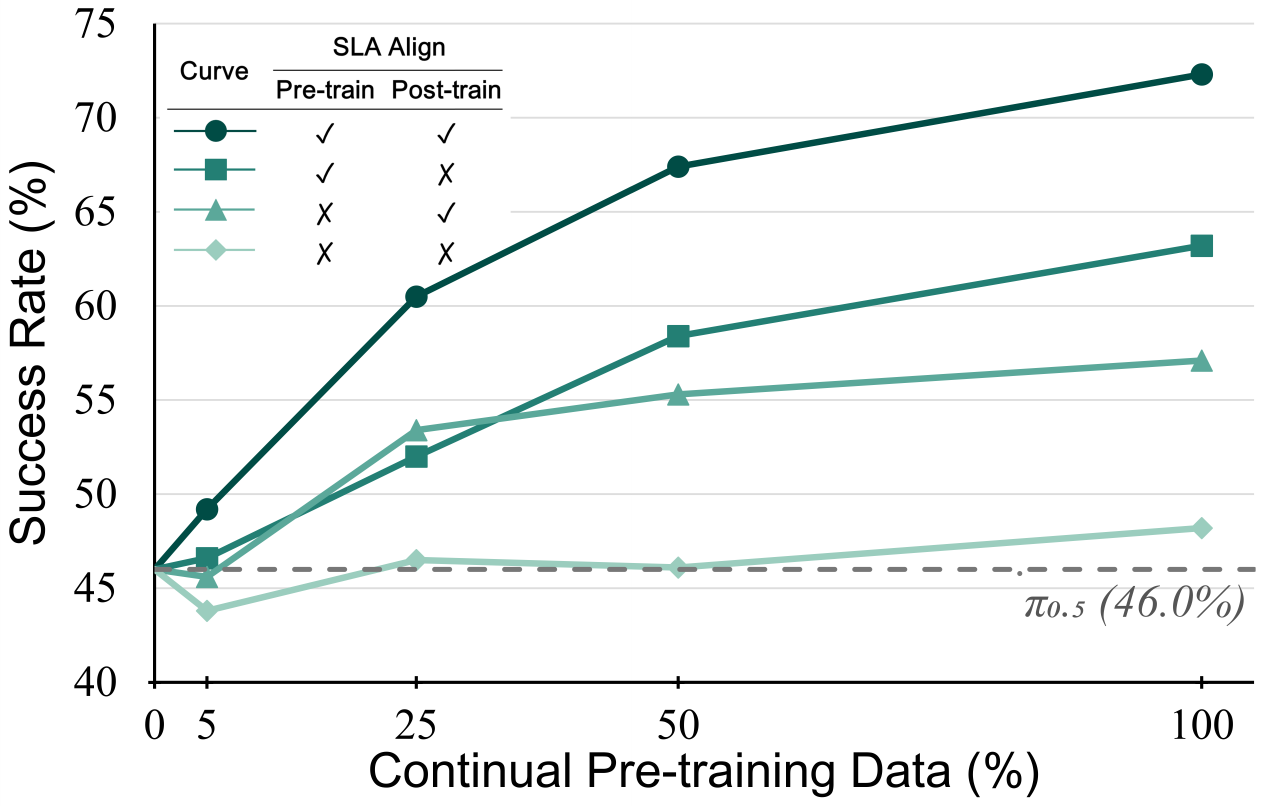}
    \caption{\textbf{Continual pre-training data scaling ablation.} \method{} alignment is independently enabled or disabled during continual pre-training and post-training. All curves start at the shared $\pi_{0.5}$ baseline. Experiments are conducted under RoboTwin Clean2Random settings to evaluate success rates (\%, $\uparrow$).}
    \label{fig:scaling}
\end{figure}

A key question for continual pre-training is whether additional training data translate into better downstream performance. To examine this relationship, we sample demonstrations within each task of the full pre-training mixture to construct subsets containing 5\%, 25\%, 50\%, and 100\% of the original data. For each subset, we perform continual pre-training with and without \method{} alignment to obtain separate checkpoints. We then independently enable or disable alignment during post-training to examine how spatial supervision in the two stages interacts, yielding the four configurations shown in \Cref{tab:scaling} and \Cref{fig:scaling}.

We report RoboTwin 2.0 Clean2Random SR averaged over all 50 tasks. With \method{} alignment in both stages, SR increases from 49.2\% at 5\% data to 72.3\% at full data, a gain of 23.1 percentage points. This configuration performs best at every tested proportion. Without alignment in either stage, SR changes from 43.8\% to 48.2\%, with a small decline between 25\% and 50\% data. Increasing data alone therefore provides a much smaller benefit in this comparison.

The two single-stage variants further clarify the role of spatial supervision. At full data, \method{} alignment only during continual pre-training reaches 63.2\%, compared with 57.1\% when applied only during post-training. Applying alignment in both stages improves on these variants by 9.1 and 15.2 percentage points, respectively. Overall, the consistently stronger joint configuration supports using spatial supervision both to shape the pretrained representation and to guide its downstream adaptation.


\subsubsection{Adaptive Layer Selection}

\begin{table}[!htbp]
    \centering
    \caption{\textbf{Adaptive layer selection ablation.} Fixed variants use visual tokens from a single VLA transformer layer for alignment; Adaptive variants use the learned fusion mechanism. 
    Experiments are conducted on RoboTwin to evaluate success rates (\%, $\uparrow$).}
    \label{tab:adaptive_layer_selection}
    \begin{tabular}{cccc}
\toprule
Alignment Layer & Clean2Clean & Clean2Random & Average \\
\midrule
1 & 60.0 & 43.4 & 51.7 \\
6 & 63.0 & 45.0 & 54.0 \\
12 & \underline{84.0} & \underline{67.3} & \underline{75.7} \\
18 & 78.0 & 64.6 & 71.3 \\
 Adaptive & \textbf{86.0} & \textbf{72.3} & \textbf{79.2} \\
\bottomrule
    \end{tabular}
\end{table}

A fixed alignment layer requires manual selection and may not suit different data domains. We therefore introduce the adaptive layer selection mechanism to learn each layer's contribution to spatial supervision, as explained in \Cref{sec:adaptive_layers}. Specifically, our PaliGemma VLM~\citep{paligemma} has 18 transformer layers in its language backbone. We compare single-layer alignment at layers 1, 6, 12, and 18 with adaptive fusion across all 18 layers. In \Cref{tab:adaptive_layer_selection}, fixed-layer Clean2Random success rates range from 43.4\% to 67.3\%. Layer 12 performs best, while layer 18 drops to 64.6\%, showing that deeper alignment is not necessarily better. Adaptive fusion achieves 72.3\% Clean2Random success rates and 79.2\% average success rates, exceeding layer 12 by 5.0 and 3.5 percentage points, respectively. This result supports learning the contributions of multiple layers instead of committing to a single layer depth. The improvement is consistent with the motivation for token- and channel-wise gating: allowing the spatial alignment signal to draw on features from different representation depths.

\subsubsection{State Dropout}

\begin{table}[!htbp]
    \centering
    \caption{\textbf{State dropout ablation.} Experiments are conducted on RoboTwin to evaluate success rates (\%, $\uparrow$).}
    \label{tab:state_dropout}
    \begin{tabular}{cccc}
\toprule
\makecell{State Dropout\\Probability} & Clean2Clean & Clean2Random & Average \\
\midrule
0\% & 55.5 & 20.7 & 38.1 \\
25\% & \textbf{86.0} & \textbf{72.3} & \textbf{79.2} \\
50\% & 66.5 & 41.3 & 53.9 \\
75\% & \underline{72.1} & \underline{51.6} & \underline{61.9} \\
100\% & 62.4 & 50.9 & 56.7 \\
\bottomrule
    \end{tabular}
\end{table}

State dropout is intended to discourage shortcuts based on associations between robot states and demonstrated trajectories, encouraging the policy to encode scene geometry and use visual observations for action generation. To assess its effect on manipulation performance, we vary the state dropout probability during post-training, starting from the same continually pretrained checkpoint and evaluating all 50 RoboTwin tasks. As shown in \Cref{tab:state_dropout}, 25\% dropout achieves the best average SR of 79.2\%, compared with 38.1\% without dropout. It also yields the highest Clean2Clean and Clean2Random results. Larger dropout probabilities perform less well: even 75\%, the next-best setting, reaches only 61.9\% on average. The response is not monotonic, and removing state inputs from every training sample is also suboptimal. These observations favor partial state dropout, consistent with a balance between encouraging reliance on visual evidence and preserving access to proprioceptive information during training.

\subsubsection{Rotation Representation}

\begin{table}[!htbp]
    \centering
    \caption{\textbf{Rotation representation ablation.} All variants compute relative rotations through $\mathrm{SO}(3)$ composition rather than direct subtraction to avoid wrap-induced outliers, as detailed in \Cref{eq:relative_action}. Experiments are conducted on RoboTwin to evaluate success rates (\%, $\uparrow$).}
    \label{tab:rotation_representation}
    \begin{tabular}{cccc}
\toprule
Rotation Representation & Clean2Clean & Clean2Random & Average \\
\midrule
Euler Angle & 71.6 & 59.3 & 65.5 \\
Rot6D & 66.9 & 58.8 & 62.9 \\
Quaternion & \underline{81.1} & \underline{67.5} & \underline{74.3} \\
Rotation Vector & \textbf{86.0} & \textbf{72.3} & \textbf{79.2} \\
\bottomrule
    \end{tabular}
\end{table}

Because our action space is defined in terms of end-effector poses, the rotation representation determines how orientation changes are encoded as action targets. We compare four rotation representations during post-training from the same continually pretrained checkpoint on all 50 RoboTwin tasks. \Cref{tab:rotation_representation} shows that rotation vectors achieve the highest average SR of 79.2\%, followed by quaternions at 74.3\%, Euler angles at 65.5\%, and Rot6D at 62.9\%. Rotation vectors also achieve the best results in both Clean2Clean and Clean2Random settings, supporting their use for downstream action learning. All four variants compute relative rotations through $\mathrm{SO}(3)$ composition, following \cref{eq:relative_action}. The comparison therefore isolates how the composed rotation is represented for learning; it does not compare composition with direct coordinate subtraction. The results identify rotation vectors as the most effective representation within this controlled post-training setting.


\subsubsection{Camera Calibration}

\begin{table}[!htbp]
    \centering
    \caption{\textbf{Camera calibration ablation.} Calibration comprises camera intrinsics and extrinsics supplied to the OmniVGGT for spatial supervision. Experiments are conducted on RoboTwin to evaluate success rates (\%, $\uparrow$). Models are fine-tuned on personal-collected RoboTwin clean data.}
    \label{tab:camera_params}
    \setlength{\tabcolsep}{5pt}
    \begin{tabular}{cccc}
\toprule
\method{} Alignment & Clean2Clean & Clean2Random & Average \\
\midrule
\Crossmark & 36.9 & 26.7 & 31.8 \\
VGGT & \underline{50.2} & 34.8 & 42.5 \\
OmniVGGT w/o Calibration & 49.9 & \underline{35.6} & \underline{42.8} \\
OmniVGGT w/ Calibration & \textbf{54.7} & \textbf{37.3} & \textbf{46.0} \\
\bottomrule
    \end{tabular}
\end{table}

OmniVGGT, our external spatial supervision model, can incorporate camera calibration alongside RGB images to provide additional geometric constraints on its spatial representations. We evaluate the benefit of this information using our collected RoboTwin Clean demonstrations across all 50 tasks. Starting from the same continually pretrained checkpoint, we change only the spatial supervision used during post-training. In \Cref{tab:camera_params}, the average SR increases from 31.8\% without post-training alignment to 42.8\% with supervision from an RGB-only OmniVGGT teacher. Providing camera intrinsics and extrinsics to this spatial foundation model further raises the average to 46.0\%, with improvements in both Clean2Clean and Clean2Random settings. Spatial supervision thus provides the larger gain, while calibration contributes an additional 3.2 percentage points. This pattern is consistent with geometric annotations complementing the spatial information inferred from RGB observations. Calibration is supplied only to the external teacher during training and is not required as an additional policy input at deployment.

\subsection{Comparisons on Real-world Deployment}
\label{sec:real_world_deployment}



\begin{figure*}[t]
    \centering
    \includegraphics[width=\textwidth]{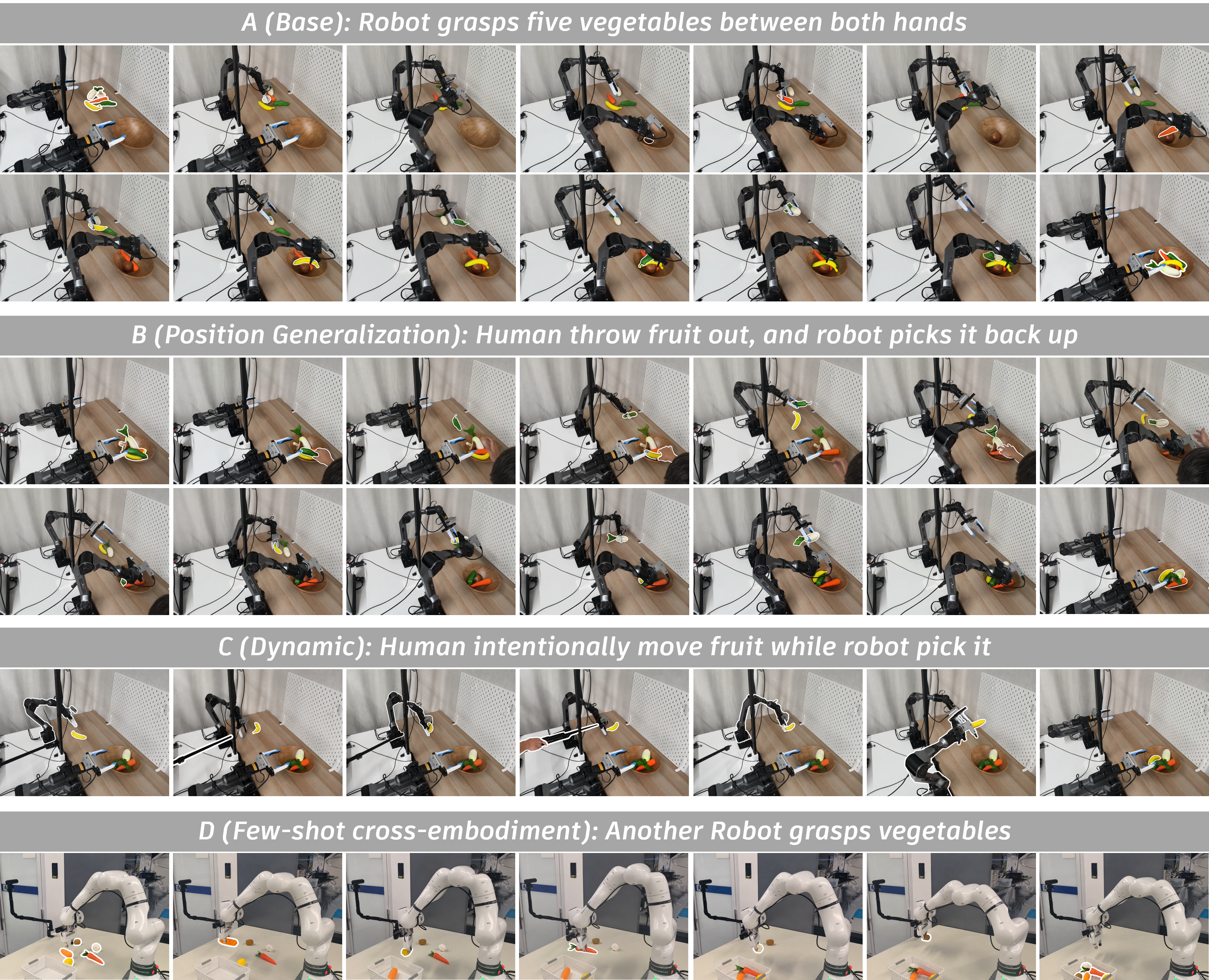}
    \caption{\textbf{Produce manipulation tasks under human perturbations and embodiment changes.} A--C conditions share a policy post-trained on 100 demonstrations collected in the A scene configuration, with no additional demonstrations of the perturbations in B or C scene. D condition evaluates a policy jointly post-trained on 100 Piper demonstrations collected in the A scene and 10 Franka demonstrations collected in the D scene.}
    \label{fig:real_world_produce}
\end{figure*}

\begin{figure*}[t]
    \centering
    \includegraphics[width=\textwidth]{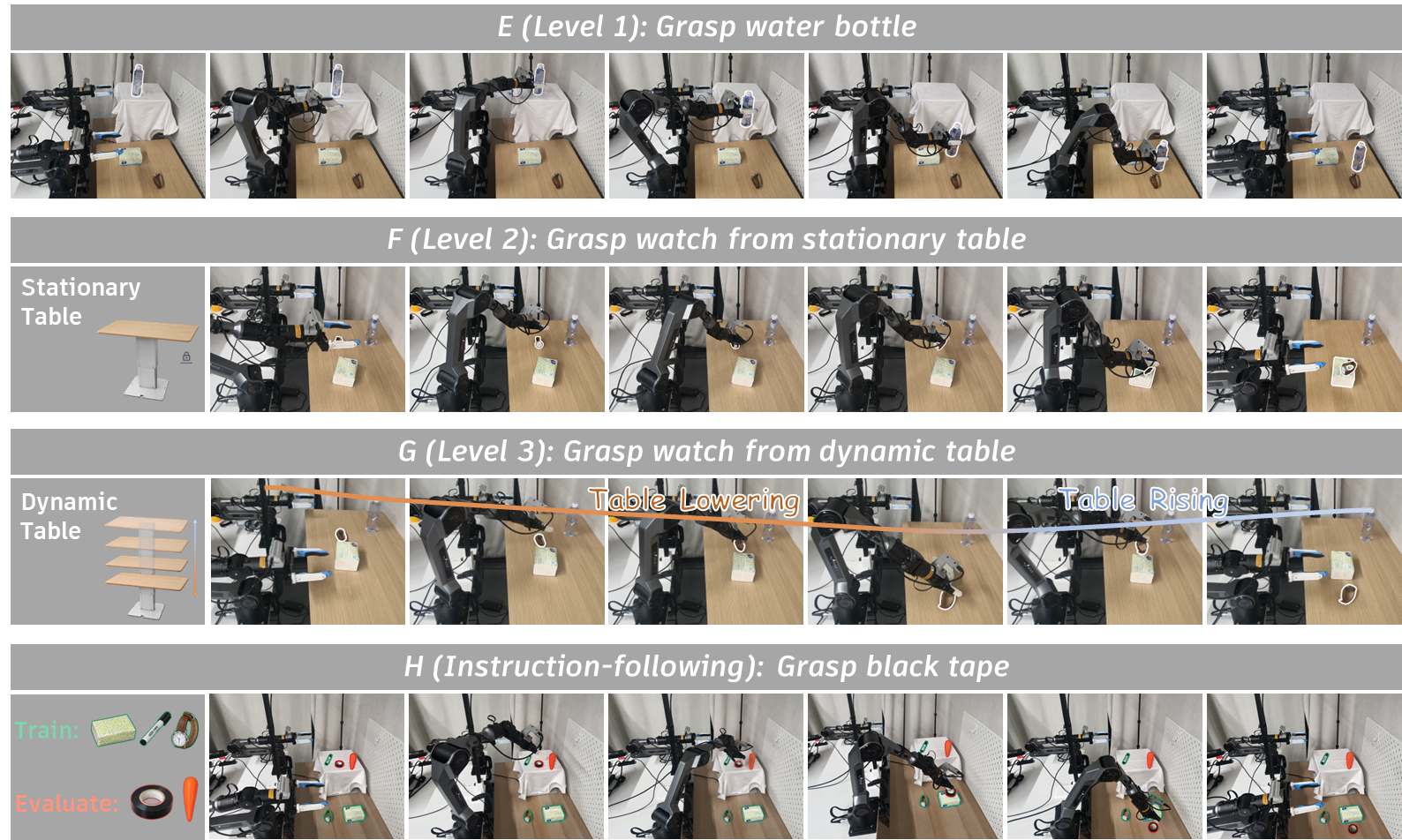}
    \caption{\textbf{Tabletop manipulation tasks under changes in object, table height, and instruction target.} All four conditions share a policy post-trained on 100 demonstrations collected in the E scene configuration. The bottle, tissue box, pen, and watch are included in these demonstrations while black tape and carrot are excluded.}
    \label{fig:real_world_tabletop}
\end{figure*}

\subsubsection{Real-world Tasks}

We design a diverse set of challenging real-world tasks to evaluate bimanual coordination, spatial and dynamic generalization, few-shot cross-embodiment adaptation, and instruction following. The training and evaluation protocols are specified in \Cref{sec:exp_setup}.

For the produce tasks in \Cref{fig:real_world_produce}, we post-train a single policy on 100 demonstrations of the base scene A and evaluates it in A--C. \textit{A (Base) Robot grasps five vegetables between both hands:} the robot passes five objects between its arms and places them into a bowl, testing bimanual coordination. \textit{B (Position Generalization) Human throw fruit out, and robot picks it back up:} we tests recovery after a person throws objects from the bowl onto the table, and the policy must retrieve them and repeat handover and placement. \textit{C (Dynamic) Human intentionally move fruit while robot pick it:} we tests responsiveness to target motion during grasping, while still requiring subsequent handover and placement. \textit{D (Few-shot cross-embodiment) Another Robot grasps vegetables:} To examine adaptation with limited target-robot data, we use a separate policy jointly post-trained on the 100 Piper demonstrations and 10 additional Franka demonstrations, and deploy it to a single Franka arm to place five objects into a basket.

For the tabletop tasks in \Cref{fig:real_world_tabletop}, we post-train one policy on 100 demonstrations collected in the E scene configuration and evaluates it across E--H. \textit{E (Level 1) Grasp water bottle:} we provide a reference for basic manipulation: the robot grasps a bottle and stably places it on the destination table. \textit{F (Level 2) Grasp watch from stationary table:} we test adaptation to a different static table height by transferring a watch that is already represented in the training data. \textit{G (Level 3) Grasp watch from dynamic table:} we further test responsiveness to changing target height as the supporting table rises and lowers during grasping, without additional table-motion demonstrations. \textit{H (Instruction-following) Grasp black tape:} Finally, we retain the E table configuration and tests instruction following with objects absent from downstream training: the robot must select and transfer black tape as instructed while ignoring a carrot distractor.

\subsubsection{Results}

\begin{table*}[!t]
    \centering
    \caption{\textbf{Real-world deployment comparisons.} Each method is evaluated on 50 separately scored object instances per task to report the object-level success rates (\%, $\uparrow$). Average is the arithmetic mean over A--H. \textbf{Bold} and \underline{underline} denote the best and second-best displayed values.}
    \label{tab:real_world_deployment}
    \footnotesize
    \setlength{\tabcolsep}{3pt}
    \begin{tabularx}{\textwidth}{l*{4}{Y}}
        \toprule
        Method & \makecell{A: Pass vegetables\\between hands} & \makecell{B: Grasp fruit\\thrown by a human} & \makecell{C: Grasp fruit\\moved by a human} & \makecell{D: Another robot\\grasps vegetables} \\
        \midrule
        OpenVLA-OFT~\citep{openvla-oft} & 62 & 38 & 30 & 36 \\
        SpatialVLA~\citep{spatialvla} & 70 & 46 & 46 & 44 \\
        $\pi_{0.5}$~\citep{pi05} & 74 & 52 & 42 & 48 \\
        \midrule
        \rowcolor{black!6} Spatial Forcing & \underline{84} & \underline{66} & \underline{58} & \underline{62} \\
        \rowcolor{black!6} \method{} (Ours) & \textbf{94} & \textbf{82} & \textbf{76} & \textbf{78} \\
        \bottomrule
    \end{tabularx}
    \par\vspace{2pt}\noindent
    \begin{tabularx}{\textwidth}{l*{4}{Y}c}
        Method & \makecell{E: Grasp\\water bottle} & \makecell{F: Grasp watch\\from static table} & \makecell{G: Grasp watch\\from moving table} & \makecell{H: Grasp unseen\\black tape} & Average \\
        \midrule
        OpenVLA-OFT~\citep{openvla-oft} & 82 & 50 & 28 & 42 & 46.0 \\
        SpatialVLA~\citep{spatialvla} & 88 & 58 & 38 & 38 & 53.5 \\
        $\pi_{0.5}$~\citep{pi05} & 90 & \underline{68} & 44 & \underline{60} & 59.8 \\
        \midrule
        \rowcolor{black!6} Spatial Forcing & \underline{94} & 64 & \underline{60} & 56 & \underline{68.0} \\
        \rowcolor{black!6} \method{} (Ours) & \textbf{98} & \textbf{84} & \textbf{78} & \textbf{80} & \textbf{83.8} \\
        \bottomrule
    \end{tabularx}
\end{table*}

We compare OpenVLA-OFT~\citep{openvla-oft}, SpatialVLA~\citep{spatialvla}, $\pi_{0.5}$~\citep{pi05}, Spatial Forcing, and \method{}. Spatial Forcing applies spatial supervision only during post-training from $\pi_{0.5}$, whereas \method{} combines spatially supervised continual pre-training with downstream post-training. As shown in \Cref{tab:real_world_deployment}, \method{} achieves the highest success rate across all tasks while Spatial Forcing ranks second on most tasks, which supports the effectiveness of our approach. The task-specific discussion below examines several complementary capabilities of our model that are essential for real-world applications.


\textit{A: Bimanual coordination.} The success rates are 74\% for $\pi_{0.5}$, 84\% for Spatial Forcing, and 94\% for \method{}, corresponding to gains of 10 and 20 percentage points over the base policy. These gains support the value of spatial supervision for the complete grasp--handover--placement sequence, with the full \method{} recipe providing an additional benefit over post-training-only Spatial Forcing.

\textit{B: Recovery from positional changes.} The success rates increase from 52\% with $\pi_{0.5}$ to 66\% with Spatial Forcing and 82\% with \method{}. Relative to A, the corresponding reductions are 22, 18, and 12 points. A smaller reduction for \method{} indicates more robust reuse of the learned manipulation skill after object relocation. Because A and B share a trained policy, this comparison concerns generalization to the intervention rather than fitting additional recovery demonstrations.

\textit{C: Response to moving targets.} The success rates are 42\%, 58\%, and 76\% for $\pi_{0.5}$, Spatial Forcing, and \method{}, respectively, placing \method{} 18 points above Spatial Forcing. This advantage is consistent with more effective use of current visual observations when the target moves during grasping, while retaining the ability to finish the subsequent handover and placement.

\textit{D: Adaptation with limited target-embodiment data.} Under the training budget of 100 Piper and 10 Franka demonstrations, the success rates are 48\% for $\pi_{0.5}$, 62\% for Spatial Forcing, and 78\% for \method{}. Such a result supports the usefulness of spatial supervision when adapting manipulation across embodiments with limited target domain data. The 16-point margin over Spatial Forcing further favors the complete \method{} recipe in this joint-training setting.

\textit{E: Execution in the training scene configuration.} The success rates of 90\%, 94\%, and 98\% means that both $\pi_{0.5}$ and our policies are near the upper end of the metric. The 4-point margin between \method{} and Spatial Forcing corresponds to two additional successful object instances out of 50. Such closely spaced values would establish basic manipulation competence while leaving limited headroom for improvement, making the changed conditions in F--H more informative about generalization.

\textit{F: Adaptation to a different table height.} The success rates of Spatial Forcing is 64\%, below the 68\% of $\pi_{0.5}$, while \method{} reaches 84\%. This result indicates that the benefits of post-training-only spatial supervision are not uniform across conditions. The 20-point margin over Spatial Forcing favors the complete recipe under the table height changes.

\textit{G: Manipulation during table motion.} The success rates are 44\% for $\pi_{0.5}$, 60\% for Spatial Forcing, and 78\% for \method{}. Compared with F, the reductions are 24, 4, and 6 points, respectively. The result means that both spatially supervised policies retain more abilities of their static-condition performance, while \method{} maintains the highest absolute success rate. This pattern supports spatial supervision under changing geometry and the additional benefit of the full recipe, suggesting that our methods has the powerful static-to-dynamic fitness.

\textit{H: Instruction-guided manipulation of an unseen object.} The success rates are 60\% for $\pi_{0.5}$, 56\% for Spatial Forcing, and 80\% for \method{}. A 24-point gain over Spatial Forcing supports the ability to follow the specified target while transferring grasp-and-place behavior to an object excluded from downstream training. This comparison attributes the gains of our method to language-following and geometric reasoning abilities.

Taken together, these results supports the significance of our \method{} training approach across spatial perturbations, embodiment adaptation, and language-guided object manipulation. our \method supervision uses auxiliary components only during training, retaining the original policy backbone and action-generation architecture for deployment. This compatibility enables continued training without adding policy computation at inference.

\section{Conclusion}
\label{sec:con}
In this work, we presented \method{}, a Spatial-Language-Action Model that develops transferable spatial understanding through continual pre-training and downstream post-training. By aligning intermediate visual representations with a frozen 3D perception model, our approach integrates geometric supervision into action learning while directly reusing pretrained VLA weights. A unified robot-data corpus and adaptive layer selection support training across heterogeneous embodiments and scenes, while available depth and calibration annotations enrich the spatial supervision. The teacher and alignment modules are required only during training. Extensive evaluations demonstrate state-of-the-art performance on RoboDojo-Real, RoboTwin 2.0 under the Clean2Random setting, RoboCasa-Panda, and LIBERO-Plus, alongside improved spatial and dynamic generalization, downstream data efficiency, and few-shot cross-embodiment adaptation. Controlled experiments further show that applying spatial alignment during both training stages consistently outperforms either single-stage alternative across the tested data scales. These findings support developing spatial understanding during continual pre-training and reinforcing it during downstream adaptation, offering a practical path toward spatially informed robotic foundation models.

\addcontentsline{toc}{section}{References}
\bibliographystyle{plainnat}
\bibliography{main}

@article{robotwin-leaderboard,
  title={Robotwin 2.0: A scalable data generator and benchmark with strong domain randomization for robust bimanual robotic manipulation},
  author={Chen, Tianxing and Chen, Zanxin and Chen, Baijun and Cai, Zijian and Liu, Yibin and Li, Zixuan and Liang, Qiwei and Lin, Xianliang and Ge, Yiheng and Gu, Zhenyu and others},
  journal={arXiv preprint arXiv:2506.18088},
  year={2025}
}

@article{robodojo-leaderboard,
  title={RoboDojo: A unified sim-and-real benchmark for comprehensive evaluation of generalist robot manipulation policies},
  author={Chen, Tianxing and Chen, Yue and Li, Zixuan and Tang, Junyuan and Su, Kailun and Lu, Haoran and Wan, Weijie and Chen, Baijun and Liu, Songling and Yan, Haowen and others},
  journal={arXiv preprint arXiv:2607.04434},
  year={2026}
}

@inproceedings{shi2025memoryvla,
  title={Memoryvla: Perceptual-cognitive memory in vision-language-action models for robotic manipulation},
  author={Shi, Hao and Xie, Bin and Liu, Yingfei and Sun, Lin and Liu, Fengrong and Wang, Tiancai and Zhou, Erjin and Fan, Haoqiang and Zhang, Xiangyu and Huang, Gao},
  booktitle={ICLR},
  year={2026}
}

@article{shi2026memoryvla++,
  title={MemoryVLA++: Temporal Modeling via Memory and Imagination in Vision-Language-Action Models},
  author={Shi, Hao and Li, Weiye and Xie, Bin and Wang, Yulin and Zhou, Renping and Wang, Tiancai and Zhang, Xiangyu and Luo, Ping and Huang, Gao},
  journal={arXiv preprint arXiv:2606.09827},
  year={2026}
}

@article{zeng2026kemo,
  title={KEMO: Event-Driven Keyframe Memory for Long-Horizon Robot Manipulation with VLA Policies},
  author={Zeng, Yihan and Ye, Minghao and Chen, Yiyuan and Shentu, Yide and Wu, Philipp and Yan, Zike and Li, Zhongyu},
  journal={arXiv preprint arXiv:2606.23589},
  year={2026}
}

@article{li2026remem,
  title={Remem-vla: Empowering vision-language-action model with memory via dual-level recurrent queries},
  author={Li, Hang and Shen, Fengyi and Chen, Dong and Yang, Liudi and Wang, Xudong and Shi, Jinkui and Bing, Zhenshan and Liu, Ziyuan and Knoll, Alois},
  journal={arXiv preprint arXiv:2603.12942},
  year={2026}
}

@article{openvla,
  title={Openvla: An open-source vision-language-action model},
  author={Kim, Moo Jin and Pertsch, Karl and Karamcheti, Siddharth and Xiao, Ted and Balakrishna, Ashwin and Nair, Suraj and Rafailov, Rafael and Foster, Ethan and Lam, Grace and Sanketi, Pannag and others},
  journal={arXiv preprint arXiv:2406.09246},
  year={2024}
}

@inproceedings{pi0,
  title={$\pi_0$: A vision-language-action flow model for general robot control},
  author={Black, Kevin and Brown, Noah and Driess, Danny and Esmail, Adnan and Equi, Michael and Finn, Chelsea and Fusai, Niccolo and Groom, Lachy and Hausman, Karol and Ichter, Brian and others},
  booktitle={RSS},
  year={2025}
}

@inproceedings{pi05,
  title={$\pi_{0.5} $: a Vision-Language-Action Model with Open-World Generalization},
  author={Physical Intelligence},
  booktitle={CoRL},
  year={2025}
}

@article{Rt-2,
  title={Rt-2: Vision-language-action models transfer web knowledge to robotic control},
  author={Brohan, Anthony and Brown, Noah and Carbajal, Justice and Chebotar, Yevgen and Chen, Xi and Choromanski, Krzysztof and Ding, Tianli and Driess, Danny and Dubey, Avinava and Finn, Chelsea and others},
  journal={arXiv preprint arXiv:2307.15818},
  year={2023}
}

@article{Rt-h,
  title={Rt-h: Action hierarchies using language},
  author={Belkhale, Suneel and Ding, Tianli and Xiao, Ted and Sermanet, Pierre and Vuong, Quon and Tompson, Jonathan and Chebotar, Yevgen and Dwibedi, Debidatta and Sadigh, Dorsa},
  journal={arXiv preprint arXiv:2403.01823},
  year={2024}
}

@inproceedings{siglip,
  title={Sigmoid loss for language image pre-training},
  author={Zhai, Xiaohua and Mustafa, Basil and Kolesnikov, Alexander and Beyer, Lucas},
  booktitle={Proceedings of the IEEE/CVF international conference on computer vision},
  pages={11975--11986},
  year={2023}
}

@article{dinov2,
  title={Dinov2: Learning robust visual features without supervision},
  author={Oquab, Maxime and Darcet, Timoth{\'e}e and Moutakanni, Th{\'e}o and Vo, Huy and Szafraniec, Marc and Khalidov, Vasil and Fernandez, Pierre and Haziza, Daniel and Massa, Francisco and El-Nouby, Alaaeldin and others},
  journal={arXiv preprint arXiv:2304.07193},
  year={2023}
}

@inproceedings{spatialvla,
  title={SpatialVLA: Exploring Spatial Representations for Visual-Language-Action Model},
  author={Qu, Delin and Song, Haoming and Chen, Qizhi and Yao, Yuanqi and Ye, Xinyi and Ding, Yan and Wang, Zhigang and Gu, JiaYuan and Zhao, Bin and Wang, Dong and others},
  booktitle={RSS},
  year={2025}
}

@article{univla,
  title={Learning to Act Anywhere with Task-centric Latent Actions},
  author={Bu, Qingwen and Yang, Yanting and Cai, Jisong and Gao, Shenyuan and Ren, Guanghui and Yao, Maoqing and Luo, Ping and Li, Hongyang},
  journal={arXiv preprint arXiv:2502.14420},
  year={2025}
}

@article{openvla-oft,
  title={Fine-tuning vision-language-action models: Optimizing speed and success},
  author={Kim, Moo Jin and Finn, Chelsea and Liang, Percy},
  journal={arXiv preprint arXiv:2502.19645},
  year={2025}
}

@article{bjorck2025gr00t,
  title={Gr00t n1: An open foundation model for generalist humanoid robots},
  author={Bjorck, Johan and Casta{\~n}eda, Fernando and Cherniadev, Nikita and Da, Xingye and Ding, Runyu and Fan, Linxi and Fang, Yu and Fox, Dieter and Hu, Fengyuan and Huang, Spencer and others},
  journal={arXiv preprint arXiv:2503.14734},
  year={2025}
}

@article{li2025UVA,
  title={Unified video action model},
  author={Li, Shuang and Gao, Yihuai and Sadigh, Dorsa and Song, Shuran},
  journal={arXiv preprint arXiv:2503.00200},
  year={2025}
}

@article{liu2024rdt,
  title={RDT-1B: a Diffusion Foundation Model for Bimanual Manipulation},
  author={Liu, Songming and Wu, Lingxuan and Li, Bangguo and Tan, Hengkai and Chen, Huayu and Wang, Zhengyi and Xu, Ke and Su, Hang and Zhu, Jun},
  journal={arXiv preprint arXiv:2410.07864},
  year={2024}
}

@article{zhang2025falcon,
  title={Falcon: Resolving visual redundancy and fragmentation in high-resolution multimodal large language models via visual registers},
  author={Zhang, Renshan and Shao, Rui and Chen, Gongwei and Zhang, Miao and Zhou, Kaiwen and Guan, Weili and Nie, Liqiang},
  journal={arXiv preprint arXiv:2501.16297},
  year={2025}
}

@inproceedings{song2026reconvla,
  title={Reconvla: Reconstructive vision-language-action model as effective robot perceiver},
  author={Song, Wenxuan and Zhou, Ziyang and Zhao, Han and Chen, Jiayi and Ding, Pengxiang and Yan, Haodong and Huang, Yuxin and Tang, Feilong and Wang, Donglin and Li, Haoang},
  booktitle={AAAI},
  year={2026}
}

@article{luo2026being,
  title={Being-H0. 7: A latent world-action model from egocentric videos},
  author={Luo, Hao and Zhang, Wanpeng and Feng, Yicheng and Zheng, Sipeng and Xu, Haiweng and Xu, Chaoyi and Xi, Ziheng and Fu, Yuhui and Lu, Zongqing},
  journal={arXiv preprint arXiv:2605.00078},
  year={2026}
}

@article{ye2026world,
  title={World action models are zero-shot policies},
  author={Ye, Seonghyeon and Ge, Yunhao and Zheng, Kaiyuan and Gao, Shenyuan and Yu, Sihyun and Kurian, George and Indupuru, Suneel and Tan, You Liang and Zhu, Chuning and Xiang, Jiannan and others},
  journal={arXiv preprint arXiv:2602.15922},
  year={2026}
}

@article{jang2025dreamgen,
  title={Dreamgen: Unlocking generalization in robot learning through video world models},
  author={Jang, Joel and Ye, Seonghyeon and Lin, Zongyu and Xiang, Jiannan and Bjorck, Johan and Fang, Yu and Hu, Fengyuan and Huang, Spencer and Kundalia, Kaushil and Lin, Yen-Chen and others},
  journal={arXiv preprint arXiv:2505.12705},
  year={2025}
}

@article{sridhar2025memer,
  title={Memer: Scaling up memory for robot control via experience retrieval},
  author={Sridhar, Ajay and Pan, Jennifer and Sharma, Satvik and Finn, Chelsea},
  journal={arXiv preprint arXiv:2510.20328},
  year={2025}
}

@article{li2025spatial,
  title={Spatial forcing: Implicit spatial representation alignment for vision-language-action model},
  author={Li, Fuhao and Song, Wenxuan and Zhao, Han and Wang, Jingbo and Ding, Pengxiang and Wang, Donglin and Zeng, Long and Li, Haoang},
  journal={arXiv preprint arXiv:2510.12276},
  year={2025}
}

@article{fei2025libero,
  title={Libero-plus: In-depth robustness analysis of vision-language-action models},
  author={Fei, Senyu and Wang, Siyin and Shi, Junhao and Dai, Zihao and Cai, Jikun and Qian, Pengfang and Ji, Li and He, Xinzhe and Zhang, Shiduo and Fei, Zhaoye and others},
  journal={arXiv preprint arXiv:2510.13626},
  year={2025}
}

@article{kim2026cosmos,
  title={Cosmos policy: Fine-tuning video models for visuomotor control and planning},
  author={Kim, Moo Jin and Gao, Yihuai and Lin, Tsung-Yi and Lin, Yen-Chen and Ge, Yunhao and Lam, Grace and Liang, Percy and Song, Shuran and Liu, Ming-Yu and Finn, Chelsea and others},
  journal={arXiv preprint arXiv:2601.16163},
  year={2026}
}

@article{geovla,
  title={Geovla: Empowering 3d representations in vision-language-action models},
  author={Sun, Lin and Xie, Bin and Liu, Yingfei and Shi, Hao and Wang, Tiancai and Cao, Jiale},
  journal={arXiv preprint arXiv:2508.09071},
  year={2025}
}

@article{zheng2025x,
  title={X-vla: Soft-prompted transformer as scalable cross-embodiment vision-language-action model},
  author={Zheng, Jinliang and Li, Jianxiong and Wang, Zhihao and Liu, Dongxiu and Kang, Xirui and Feng, Yuchun and Zheng, Yinan and Zou, Jiayin and Chen, Yilun and Zeng, Jia and others},
  journal={arXiv preprint arXiv:2510.10274},
  year={2025}
}

@article{yuan2026qwen,
  title={Qwen-robotmanip technical report: Alignment unlocks scale for robotic manipulation foundation models},
  author={Yuan, Haoqi and Liang, Zhixuan and Chen, Anzhe and Wang, Ye and Li, Haoyang and Lin, Pei and Huang, Yiyang and Lei, Zixing and Zhang, Tong and Zhang, Jiazhao and others},
  journal={arXiv preprint arXiv:2606.17846},
  year={2026}
}

@article{chen2025robotwin,
  title={Robotwin 2.0: A scalable data generator and benchmark with strong domain randomization for robust bimanual robotic manipulation},
  author={Chen, Tianxing and Chen, Zanxin and Chen, Baijun and Cai, Zijian and Liu, Yibin and Li, Zixuan and Liang, Qiwei and Lin, Xianliang and Ge, Yiheng and Gu, Zhenyu and others},
  journal={arXiv preprint arXiv:2506.18088},
  year={2025}
}

@article{abc130k,
  title={Scalable Behavior Cloning with Open Data, Training, and Evaluation},
  author={Allshire, Arthur and Singh, Himanshu Gaurav and Singh, Ritvik and Rashid, Adam and Choi, Hongsuk and McAllister, David and Yu, Justin and Chen, Yiyuan and Huang, Huang and Abbeel, Pieter and others},
  journal={arXiv preprint arXiv:2606.27375},
  year={2026}
}

@inproceedings{agentfounder,
  title={Scaling agents via continual pre-training},
  author={Su, Liangcai and Zhang, Zhen and Li, Guangyu and Chen, Zhuo and Wang, Chenxi and Song, Maojia and Wang, Xinyu and Li, Kuan and Wu, Jialong and Chen, Xuanzhong and others},
  booktitle={International Conference on Learning Representations},
  volume={2026},
  pages={22634--22659},
  year={2026}
}

@article{axis,
  title={AXIS: A Growable Community-Driven Data Engine for Scalable Robot Manipulation},
  author={Zhao, Mengfei and Huang, Dihong and Tang, Yikai and Li, Peihao and Yan, Mingxuan and Zhuang, Ruiqi and Huang, Yanjia and Wang, Jie and Zhai, Hai and Zhou, Tony and others},
  journal={arXiv preprint arXiv:2607.21588},
  year={2026}
}

@article{BioBERT,
  title={BioBERT: a pre-trained biomedical language representation model for biomedical text mining},
  author={Lee, Jinhyuk and Yoon, Wonjin and Kim, Sungdong and Kim, Donghyeon and Kim, Sunkyu and So, Chan Ho and Kang, Jaewoo},
  journal={Bioinformatics},
  volume={36},
  number={4},
  pages={1234--1240},
  year={2020},
  publisher={Oxford University Press}
}

@inproceedings{bridgev2,
  title={Bridgedata v2: A dataset for robot learning at scale},
  author={Walke, Homer Rich and Black, Kevin and Zhao, Tony Z and Vuong, Quan and Zheng, Chongyi and Hansen-Estruch, Philippe and He, Andre Wang and Myers, Vivek and Kim, Moo Jin and Du, Max and others},
  booktitle={Conference on robot learning},
  pages={1723--1736},
  year={2023},
  organization={PMLR}
}

@article{bridgevla,
  title={Bridgevla: Input-output alignment for efficient 3d manipulation learning with vision-language models},
  author={Li, Peiyan and Chen, Yixiang and Wu, Hongtao and Ma, Xiao and Wu, Xiangnan and Huang, Yan and Wang, Liang and Kong, Tao and Tan, Tieniu},
  journal={Advances in Neural Information Processing Systems},
  volume={38},
  pages={63635--63673},
  year={2026}
}

@article{et-vla,
  title={Embodiment Transfer Learning for Vision-Language-Action Models},
  author={Li, Chengmeng and Peng, Yaxin},
  journal={arXiv preprint arXiv:2511.01224},
  year={2025}
}

@article{FinBERT,
  title={Finbert: Financial sentiment analysis with pre-trained language models},
  author={Araci, Dogu},
  journal={arXiv preprint arXiv:1908.10063},
  year={2019}
}

@article{fractal,
  title={Rt-1: Robotics transformer for real-world control at scale},
  author={Brohan, Anthony and Brown, Noah and Carbajal, Justice and Chebotar, Yevgen and Dabis, Joseph and Finn, Chelsea and Gopalakrishnan, Keerthana and Hausman, Karol and Herzog, Alex and Hsu, Jasmine and others},
  journal={arXiv preprint arXiv:2212.06817},
  year={2022}
}

@inproceedings{galaxeaopenworld,
  title={Galaxy Open-World Dataset and G0 Dual-System VLA Model},
  author={Jiang, Tao and Yuan, Tianyuan and Liu, Yicheng and Lu, Chenhao and Cui, Jianning and Liu, Xiao and Cheng, Shuiqi and Gao, Jiyang and Xu, Huazhe and Zhao, Hang},
  booktitle={2026 IEEE International Conference on Robotics and Automation (ICRA)},
  pages={20749--20756},
  year={2026},
  organization={IEEE}
}

@article{gr2,
  title={Gr-2: A generative video-language-action model with web-scale knowledge for robot manipulation},
  author={Cheang, Chi-Lam and Chen, Guangzeng and Jing, Ya and Kong, Tao and Li, Hang and Li, Yifeng and Liu, Yuxiao and Wu, Hongtao and Xu, Jiafeng and Yang, Yichu and others},
  journal={arXiv preprint arXiv:2410.06158},
  year={2024}
}

@article{humanscale,
  title={HumanScale: Egocentric Human Video Can Outperform Real-Robot Data for Embodied Pretraining},
  author={Ma, Juncheng and Bi, Jianxin and Deng, Yufan and Zhai, Xuanran and Zhang, Kewei and Huang, Ye and Liang, Bo and Gong, Shukai and Tu, Jiankai and Tang, Xiaotian and others},
  journal={arXiv preprint arXiv:2606.20521},
  year={2026}
}

@inproceedings{llemma,
  title={Llemma: An open language model for mathematics},
  author={Azerbayev, Zhangir and Schoelkopf, Hailey and Paster, Keiran and Dos Santos, Marco and McAleer, Stephen and Jiang, Qiaochu and Deng, Jia and Biderman, Stella R and Welleck, Sean},
  booktitle={International Conference on Learning Representations},
  volume={2024},
  pages={40622--40649},
  year={2024}
}

@article{mindvla,
  title={Mind-VLA: Instruction-Aware Spatial Representation Alignment for Vision-Language-Action Models},
  author={Ding, Xingyu and Zhao, Yuzhong and Wu, Yang and Zhao, Chunhai and Zhao, Chaoyang and Zhang, Yifan and Cheng, Jian},
  journal={arXiv preprint arXiv:2608.04633},
  year={2026}
}

@inproceedings{molmoact,
  title={Molmoact: Action reasoning models that can reason in space},
  author={Lee, Jason and Duan, Jiafei and Fang, Haoquan and Deng, Yuquan and Liu, Shuo and Li, Boyang and Fang, Bohan and Zhang, Jieyu and Wang, Yi Ru and Lee, Sangho and others},
  booktitle={2026 IEEE International Conference on Robotics and Automation (ICRA)},
  pages={2585--2592},
  year={2026},
  organization={IEEE}
}

@article{molmoact2,
  title={Molmoact2: Action reasoning models for real-world deployment},
  author={Fang, Haoquan and Duan, Jiafei and Clay, Donovan and Wang, Sam and Liu, Shuo and Huang, Weikai and Fan, Xiang and Tsai, Wei-Chuan and Chen, Shirui and Wang, Yi Ru and others},
  journal={arXiv preprint arXiv:2605.02881},
  year={2026}
}

@inproceedings{oxe,
  title={Open x-embodiment: Robotic learning datasets and rt-x models: Open x-embodiment collaboration 0},
  author={O’Neill, Abby and Rehman, Abdul and Maddukuri, Abhiram and Gupta, Abhishek and Padalkar, Abhishek and Lee, Abraham and Pooley, Acorn and Gupta, Agrim and Mandlekar, Ajay and Jain, Ajinkya and others},
  booktitle={2024 IEEE International Conference on Robotics and Automation (ICRA)},
  pages={6892--6903},
  year={2024},
  organization={IEEE}
}

@article{pointvla,
  title={Pointvla: Injecting the 3d world into vision-language-action models},
  author={Li, Chengmeng and Wen, Junjie and Peng, Yaxin and Peng, Yan and Zhu, Yichen},
  journal={IEEE Robotics and Automation Letters},
  volume={11},
  number={3},
  pages={2506--2513},
  year={2026},
  publisher={IEEE}
}

@article{posevla,
  title={PoseVLA: Universal Pose Pretraining for Generalizable Vision-Language-Action Policies},
  author={Lin, Haitao and Yu, Hanyang and Huang, Jingshun and Zhang, He and Ling, Yonggen and Tan, Ping and Xue, Xiangyang and Fu, Yanwei},
  journal={arXiv preprint arXiv:2602.19710},
  year={2026}
}

@article{rdt2,
  title={Rdt2: Exploring the scaling limit of umi data towards zero-shot cross-embodiment generalization},
  author={Liu, Songming and Li, Bangguo and Ma, Kai and Wu, Lingxuan and Tan, Hengkai and Ouyang, Xiao and Su, Hang and Zhu, Jun},
  journal={arXiv preprint arXiv:2602.03310},
  year={2026}
}

@inproceedings{rh20t,
  title={Rh20t: A comprehensive robotic dataset for learning diverse skills in one-shot},
  author={Fang, Hao-Shu and Fang, Hongjie and Tang, Zhenyu and Liu, Jirong and Wang, Chenxi and Wang, Junbo and Zhu, Haoyi and Lu, Cewu},
  booktitle={2024 IEEE International Conference on Robotics and Automation (ICRA)},
  pages={653--660},
  year={2024},
  organization={IEEE}
}

@article{robomind,
  title={Robomind: Benchmark on multi-embodiment intelligence normative data for robot manipulation},
  author={Wu, Kun and Hou, Chengkai and Liu, Jiaming and Che, Zhengping and Ju, Xiaozhu and Yang, Zhuqin and Li, Meng and Zhao, Yinuo and Xu, Zhiyuan and Yang, Guang and others},
  journal={arXiv preprint arXiv:2412.13877},
  year={2024}
}

@article{rocket,
  title={ROCKET: Residual-Oriented Multi-Layer Alignment for Spatially-Aware Vision-Language-Action Models},
  author={Sun, Guoheng and Du, Tingting and Feng, Kaixi and Luo, Chenxiang and Ding, Xingguo and Shen, Zheyu and Wang, Ziyao and He, Yexiao and Li, Ang},
  journal={arXiv preprint arXiv:2602.17951},
  year={2026}
}

@article{umi,
  title={Universal manipulation interface: In-the-wild robot teaching without in-the-wild robots},
  author={Chi, Cheng and Xu, Zhenjia and Pan, Chuer and Cousineau, Eric and Burchfiel, Benjamin and Feng, Siyuan and Tedrake, Russ and Song, Shuran},
  journal={arXiv preprint arXiv:2402.10329},
  year={2024}
}

@article{vega,
  title={VEGA: Visual Encoder Grounding Alignment for Spatially-Aware Vision-Language-Action Models},
  author={Wang, Hao and Wei, Xiaobao and He, Jingyang and Bai, Chengyu and Fan, Chun-Kai and Cao, Jiajun and Chen, Jintao and Li, Ying and Rong, Shanyu and Lu, Ming and others},
  journal={arXiv preprint arXiv:2605.10485},
  year={2026}
}

@inproceedings{vipa-vla,
  title={Spatial-aware vla pretraining through visual-physical alignment from human videos},
  author={Feng, Yicheng and Zhang, Wanpeng and Wang, Ye and Luo, Hao and Yuan, Haoqi and Zheng, Sipeng and Lu, Zongqing},
  booktitle={Proceedings of the IEEE/CVF Conference on Computer Vision and Pattern Recognition},
  pages={712--723},
  year={2026}
}

@article{xiaomi_robotics0,
  title={Xiaomi-robotics-0: An open-sourced vision-language-action model with real-time execution},
  author={Cai, Rui and Guo, Jun and He, Xinze and Jin, Piaopiao and Li, Jie and Lin, Bingxuan and Liu, Futeng and Liu, Wei and Ma, Fei and Ma, Kun and others},
  journal={arXiv preprint arXiv:2602.12684},
  year={2026}
}

@article{xiaomi_robotics1,
  title={Xiaomi-Robotics-1: Scaling Vision-Language-Action Models with over 100K Hours of Real-World Trajectories},
  author={Team, Xiaomi Robotics and Guo, Jun and Jin, Piaopiao and Li, Jason and Li, Peiyan and Li, Yingyan and Liu, Futeng and Peng, Wanli and Qin, Optimus and Su, Yifei and others},
  journal={arXiv preprint arXiv:2607.15330},
  year={2026}
}

@inproceedings{ClinicalBERT,
  title={Publicly available clinical BERT embeddings},
  author={Alsentzer, Emily and Murphy, John and Boag, William and Weng, Wei-Hung and Jindi, Di and Naumann, Tristan and McDermott, Matthew},
  booktitle={Proceedings of the 2nd clinical natural language processing workshop},
  pages={72--78},
  year={2019}
}

@inproceedings{Dont-Stop-Pretraining,
  title={Don’t stop pretraining: Adapt language models to domains and tasks},
  author={Gururangan, Suchin and Marasovi{\'c}, Ana and Swayamdipta, Swabha and Lo, Kyle and Beltagy, Iz and Downey, Doug and Smith, Noah A},
  booktitle={Proceedings of the 58th annual meeting of the association for computational linguistics},
  pages={8342--8360},
  year={2020}
}

@inproceedings{LEGAL-BERT,
  title={LEGAL-BERT: The muppets straight out of law school},
  author={Chalkidis, Ilias and Fergadiotis, Manos and Malakasiotis, Prodromos and Aletras, Nikolaos and Androutsopoulos, Ion},
  booktitle={Findings of the association for computational linguistics: EMNLP 2020},
  pages={2898--2904},
  year={2020}
}

@article{GAP,
  title={When would vision-proprioception policies fail in robotic manipulation?},
  author={Lu, Jingxian and Xia, Wenke and Wu, Yuxuan and Lu, Zhiwu and Hu, Di},
  journal={arXiv preprint arXiv:2602.12032},
  year={2026}
}

@article{state-free-policy,
  title={Do you need proprioceptive states in visuomotor policies?},
  author={Zhao, Juntu and Lu, Wenbo and Zhang, Di and Liu, Yufeng and Liang, Yushen and Zhang, Tianluo and Cao, Yifeng and Xie, Junyuan and Hu, Yingdong and Wang, Shengjie and others},
  journal={arXiv preprint arXiv:2509.18644},
  year={2025}
}

@inproceedings{vggt,
  title={{VGGT}: Visual Geometry Grounded Transformer},
  author={Wang, Jianyuan and Chen, Minghao and Karaev, Nikita and Vedaldi, Andrea and Rupprecht, Christian and Novotny, David},
  booktitle={Proceedings of the IEEE/CVF Conference on Computer Vision and Pattern Recognition},
  pages={5294--5306},
  year={2025}
}

@article{omnivggt,
  title={{OmniVGGT}: Omni-Modality Driven Visual Geometry Grounded Transformer},
  author={Peng, Haosong and Li, Hao and Dai, Yalun and Lan, Yushi and Luo, Yihang and Qi, Tianyu and Zhang, Zhengshen and Zhan, Yufeng and Zhang, Junfei and Xu, Wenchao and Liu, Ziwei},
  journal={arXiv preprint arXiv:2511.10560},
  year={2025}
}

@article{gated-attention,
  title={Gated Attention for Large Language Models: Non-linearity, Sparsity, and Attention-Sink-Free},
  author={Qiu, Zihan and Wang, Zekun and Zheng, Bo and Huang, Zeyu and Wen, Kaiyue and Yang, Songlin and Men, Rui and Yu, Le and Huang, Fei and Huang, Suozhi and Liu, Dayiheng and Zhou, Jingren and Lin, Junyang},
  journal={arXiv preprint arXiv:2505.06708},
  year={2025}
}

@article{geometry-forcing,
  title={Geometry Forcing: Marrying Video Diffusion and 3D Representation for Consistent World Modeling},
  author={Wu, Haoyu and Wu, Diankun and He, Tianyu and Guo, Junliang and Ye, Yang and Duan, Yueqi and Bian, Jiang},
  journal={arXiv preprint arXiv:2507.07982},
  year={2025}
}

@inproceedings{dpt,
  title={Vision Transformers for Dense Prediction},
  author={Ranftl, Ren{\'e} and Bochkovskiy, Alexey and Koltun, Vladlen},
  booktitle={Proceedings of the IEEE/CVF International Conference on Computer Vision},
  pages={12179--12188},
  year={2021}
}

@article{robocoin,
  title={{RoboCOIN}: An Open-Sourced Bimanual Robotic Data Collection for Integrated Manipulation},
  author={Wu, Shihan and Liu, Xuecheng and Xie, Shaoxuan and Wang, Pengwei and Li, Xinghang and Yang, Bowen and others},
  journal={arXiv preprint arXiv:2511.17441},
  year={2025}
}

@article{agibot-world,
  title={{AgiBot World Colosseo}: A Large-scale Manipulation Platform for Scalable and Intelligent Embodied Systems},
  author={{AgiBot-World-Contributors} and Bu, Qingwen and Cai, Jisong and Chen, Li and Cui, Xiuqi and Ding, Yan and others},
  journal={arXiv preprint arXiv:2503.06669},
  year={2025}
}

@article{droid,
  title={{DROID}: A Large-Scale In-The-Wild Robot Manipulation Dataset},
  author={Khazatsky, Alexander and Pertsch, Karl and Nair, Suraj and Balakrishna, Ashwin and Dasari, Sudeep and Karamcheti, Siddharth and others},
  journal={arXiv preprint arXiv:2403.12945},
  year={2024}
}

@article{interndata-a1,
  title={{InternData-A1}: Pioneering High-Fidelity Synthetic Data for Pre-training Generalist Policy},
  author={Tian, Yang and Yang, Yuyin and Xie, Yiman and Cai, Zetao and Shi, Xu and Gao, Ning and Liu, Hangxu and Jiang, Xuekun and Qiu, Zherui and Yuan, Feng and Li, Yaping and Wang, Ping and Cai, Junhao and Zeng, Jia and Dong, Hao and Pang, Jiangmiao},
  journal={arXiv preprint arXiv:2511.16651},
  year={2025}
}

@article{robocasa365,
  title={{RoboCasa365}: A Large-Scale Simulation Framework for Training and Benchmarking Generalist Robots},
  author={Nasiriany, Soroush and Nasiriany, Sepehr and Maddukuri, Abhiram and Zhu, Yuke},
  journal={arXiv preprint arXiv:2603.04356},
  year={2026}
}

@article{flow-matching,
  title={Flow matching for generative modeling},
  author={Lipman, Yaron and Chen, Ricky TQ and Ben-Hamu, Heli and Nickel, Maximilian and Le, Matt},
  journal={arXiv preprint arXiv:2210.02747},
  year={2022}
}

@inproceedings{linearprobing,
  title={Momentum contrast for unsupervised visual representation learning},
  author={He, Kaiming and Fan, Haoqi and Wu, Yuxin and Xie, Saining and Girshick, Ross},
  booktitle={2020 IEEE/CVF conference on computer vision and pattern recognition (CVPR)},
  pages={9726--9735},
  year={2020},
  organization={IEEE}
}

@article{starvla,
  title={StarVLA: A Lego-like Codebase for Vision-Language-Action Model Developing},
  author={Community, StarVLA},
  journal={arXiv preprint arXiv:2604.05014},
  year={2026}
}

@article{abot-m0,
  title={Abot-m0: Vla foundation model for robotic manipulation with action manifold learning, 2026},
  author={Yang, Yandan and Zeng, Shuang and Lin, Tong and Chang, Xinyuan and Qi, Dekang and Xiao, Junjin and Liu, Haoyun and Chen, Ronghan and Chen, Yuzhi and Huo, Dongjie and others},
  journal={arXiv preprint arXiv:2602.11236},
  year={2026}
}

@article{4d-wam,
  title={4d-wam: Infusing spatiotemporal awareness into world action models through trajectory fields},
  author={Yang, Lishan and Song, Wenxuan and Wang, Xi and Sheng, Pingyue and Fang, Zheng and Zhou, Ziyang and He, Junjie and Yan, Haodong and Chen, Jiayi and Sun, Nan and others},
  journal={arXiv preprint arXiv:2608.08023},
  year={2026}
}

@article{worldscape2,
  title={WorldScape Policy 2.0: Empowering Steerable World Action Modeling with Reasoning-Augmented Memory},
  author={Su, Haisheng and Liu, Zongdai and Jin, Xin and Dou, Haoxuan and Hu, Chengming and Li, Baorun and Liu, Zhanwang and Xu, Ruiyan and Fang, Jianjie and Zhang, Xin and others},
  journal={arXiv preprint arXiv:2607.18840},
  year={2026}
}

@article{gigabrain07,
  title={Gigabrain-0.7: Scaling embodied foundation models to emergent capabilities with a three-system architecture},
  author={Team, GigaBrain and Ye, Angen and Sun, Axiang and Jin, Can and Cheng, Chenxi and Shi, Chong and Shang, Dengke and Zhang, Dingqian and Huang, Guan and Wang, Guangqiang and others},
  journal={arXiv preprint arXiv:2608.15875},
  year={2026}
}

@article{uniwam,
  title={UniWAM: Unified World-Action Model},
  author={Chen, Jiayi and Song, Wenxuan and Wang, Jingbo and Zhou, Shuai and Gong, Xicheng and Fan, Zehua and Zhou, Ziyang and Yan, Haodong and Li, Fuhao and Yu, Qize and others},
  journal={arXiv preprint arXiv:2610.02054},
  year={2026}
}

@article{vigar,
  title={Rethinking World-Action Model for Compositional and In-Context Robotic Manipulation},
  author={Gong, Shukai and Zhai, Xuanran and Zhang, Yintianrun and Cui, Ruopeng and Huang, Ye and Fu, Yiyang and Lyu, Dexuan and Li, Chaojie and Song, Xinyi and Lin, Peiwen and others},
  journal={arXiv preprint arXiv:2610.02368},
  year={2026}
}

@article{medex,
  title={ME-Dex 1.0: Bringing Heterogeneous Tactile Sensing into World Action Modeling},
  author={Zhang, Xuancheng and Liu, Xuetao and Tang, Qianying and Wang, Jizhe and Wang, Runsheng and Cheng, Zhijing and Lin, Bochen and Wen, Haoran and Li, Ming and Zhan, Kun and others},
  journal={arXiv preprint arXiv:2609.21449},
  year={2026}
}

@article{dp-vla,
  title={A dual process vla: Efficient robotic manipulation leveraging vlm},
  author={Han, ByungOk and Kim, Jaehong and Jang, Jinhyeok},
  journal={arXiv preprint arXiv:2410.15549},
  year={2024}
}

@article{dust,
  title={Dual-stream diffusion for world-model augmented vision-language-action model},
  author={Won, John and Lee, Kyungmin and Jang, Huiwon and Kim, Dongyoung and Shin, Jinwoo},
  journal={arXiv preprint arXiv:2510.27607},
  year={2025}
}

@article{uwm,
  title={Unified world models: Coupling video and action diffusion for pretraining on large robotic datasets},
  author={Zhu, Chuning and Yu, Raymond and Feng, Siyuan and Burchfiel, Benjamin and Shah, Paarth and Gupta, Abhishek},
  journal={arXiv preprint arXiv:2504.02792},
  year={2025}
}

@article{video-policy,
  title={Video generators are robot policies},
  author={Liang, Junbang and Tokmakov, Pavel and Liu, Ruoshi and Sudhakar, Sruthi and Shah, Paarth and Ambrus, Rares and Vondrick, Carl},
  journal={arXiv preprint arXiv:2508.00795},
  year={2025}
}

@article{flare,
  title={Flare: Robot learning with implicit world modeling},
  author={Zheng, Ruijie and Wang, Jing and Reed, Scott and Bjorck, Johan and Fang, Yu and Hu, Fengyuan and Jang, Joel and Kundalia, Kaushil and Lin, Zongyu and Magne, Loic and others},
  journal={arXiv preprint arXiv:2505.15659},
  year={2025}
}

@inproceedings{hamlet,
  title={Hamlet: Switch your vision-language-action model into a history-aware policy},
  author={Koo, Myungkyu and Choi, Daewon and Kim, Taeyoung and Lee, Kyungmin and Kim, Changyeon and Seo, Younggyo and Shin, Jinwoo},
  booktitle={International Conference on Learning Representations},
  volume={2026},
  pages={101537--101558},
  year={2026}
}

@article{internvla-a1,
  title={Internvla-a1: Unifying understanding, generation and action for robotic manipulation},
  author={Cai, Junhao and Cai, Zetao and Cao, Jiafei and Chen, Yilun and He, Zeyu and Jiang, Lei and Li, Hang and Li, Hengjie and Li, Yang and Liu, Yufei and others},
  journal={arXiv preprint arXiv:2601.02456},
  year={2026}
}

@article{openwam,
  title={Openwam: An open, modular exploration towards systematic world-action model pretraining},
  author={Wang, Yuran and Huang, Siqiao and Li, Mingleyang and Zhang, Chenhao and Liang, Jiaqi and Jin, Weiyang and Chen, Yue and Chi, Xuemin and Zhou, Donghao and Yu, Qize and others},
  journal={arXiv preprint arXiv:2609.07398},
  year={2026}
}

@misc{gr00t-n15,
  author={{NVIDIA}},
  title={{GR00T N1.5}: An Improved Open Foundation Model for Generalist Humanoid Robots},
  year={2025},
  url={https://research.nvidia.com/labs/gear/gr00t-n1_5/},
}

@article{spirit-v15,
  author = {Spirit AI Team},
  title = {Spirit-v1.5: Clean Data Is the Enemy of Great Robot Foundation Models},
  journal = {Spirit AI Blog},
  year = {2026},
  note = {https://www.spirit-ai.com/en/blog/spirit-v1-5},
}

@article{robocasa,
  title={Robocasa: Large-scale simulation of everyday tasks for generalist robots},
  author={Nasiriany, Soroush and Maddukuri, Abhiram and Zhang, Lance and Parikh, Adeet and Lo, Aaron and Joshi, Abhishek and Mandlekar, Ajay and Zhu, Yuke},
  journal={arXiv preprint arXiv:2406.02523},
  year={2024}
}

@article{qwenrobotmanip,
  title={Qwen-robotmanip technical report: Alignment unlocks scale for robotic manipulation foundation models},
  author={Yuan, Haoqi and Liang, Zhixuan and Chen, Anzhe and Wang, Ye and Li, Haoyang and Lin, Pei and Huang, Yiyang and Lei, Zixing and Zhang, Tong and Zhang, Jiazhao and others},
  journal={arXiv preprint arXiv:2606.17846},
  year={2026}
}

@article{g05,
  title={G0. 5: One autoregressive stream for robot reasoning and action},
  author={Liu, Yicheng and Dong, Zibin and Ye, Baijun and Yuan, Tianyuan and Jiang, Tao and Yang, Anqi and Cao, Shicheng and Liu, Haonan and Sun, Yue and Guo, Zihan and others},
  journal={arXiv preprint arXiv:2608.11739},
  year={2026}
}

@article{paligemma,
  title={Paligemma: A versatile 3b vlm for transfer},
  author={Beyer, Lucas and Steiner, Andreas and Pinto, Andr{\'e} Susano and Kolesnikov, Alexander and Wang, Xiao and Salz, Daniel and Neumann, Maxim and Alabdulmohsin, Ibrahim and Tschannen, Michael and Bugliarello, Emanuele and others},
  journal={arXiv preprint arXiv:2407.07726},
  year={2024}
}

@article{internw0,
  title={{InternW0-$\Delta$}: A World Action Model Bridging Predictive Dynamics and Actions with 20K+ Hours of Open Data},
  author={Miao, Xingyu and Li, Zizun and Fang, Baole and Song, Kaiwen and Wang, Tenghui and Zhang, Hanxue and Wang, Yating and Li, Xudong and He, Yuping and Wei, Xueyuan and others},
  journal={arXiv preprint arXiv:2609.31394},
  year={2026}
}

@article{x-wam,
  title={Unified 4d world action modeling from video priors with asynchronous denoising},
  author={Guo, Jun and Li, Qiwei and Li, Peiyan and Chen, Zilong and Sun, Nan and Su, Yifei and Wang, Heyun and Zhang, Yuan and Li, Xinghang and Liu, Huaping},
  journal={arXiv preprint arXiv:2604.26694},
  year={2026}
}

@article{discreteforcing,
  title={Discrete Forcing: Infusing Discrete Guidance into Continuous Denoising for Few-Step Action Experts},
  author={Wang, Jingbo and Song, Wenxuan and Yu, Wenhao and Zhao, Han and Wang, Xi and Chen, Jiayi and Wang, Donglin and Wang, Yan and Li, Haoang},
  journal={arXiv preprint arXiv:2609.39526},
  year={2026}
}

@article{vpp2,
  title={Video Prediction Policy 2: Predict Better, Act Better},
  author={Guo, Yanjiang and Yan, Haodong and Zhong, Zhide and Zhang, Zhongru and Yang, Qingyuan and Lu, Qingzhou and Chen, Xiaoyu and Wang, Yen-Jen and Deng, Shuying and Yang, Chenghan and others},
  journal={arXiv preprint arXiv:2610.10270},
  year={2026}
}

@article{groundingpi,
  title={GroundingPI: A Grounding Foundation Model towards Physical Intelligence with Visual Primitives},
  author={Yu, Qize and Fan, Lianrui and Chen, Boyu and Liang, Jiaqi and Ding, Xini and Chen, Yue and Song, Zetian and Wang, Yuran and Zou, Yi and Wang, Kaixuan and others},
  journal={arXiv preprint arXiv:2609.39601},
  year={2026}
}

@article{simplememvla,
  title={SimpleMemVLA: A Simple but Effective Native-Video Memory for Vision-Language-Action Models},
  author={Yin, Cheng and Xu, Wang and Yang, Junpeng and Tam, Sikyuen and Liu, Hanyu and Yao, Yuan and Zeng, Xiangrui and Cui, Junbo and Wang, Yequan and Yin, Zhouping and others},
  journal={arXiv preprint arXiv:2609.05533},
  year={2026}
}

@article{kinrt,
  title={Route by Kinematics, Act by Observation: Kinematics-Supervised Expert Routing in MoE-Augmented VLA},
  author={Yang, Tianhang and Zheng, Yanze and Wang, Junjie and Kou, Wei-Bin and Li, Ruotong and Yang, Yujiu},
  journal={arXiv preprint arXiv:2607.26807},
  year={2026}
}

@article{cogwam,
  title={CogWAM: Aligning Semantic Cognition with World Action Modeling via Event-Driven Interfaces},
  author={Wang, Sen and Liu, Liu and Wang, Xinjiang and Chen, Zequn and Jiang, Haoyi and Ding, Taojun and Xiao, Tingyang and Su, Zhizhong and Wang, Jie and Zhou, Sanping},
  journal={arXiv preprint arXiv:2609.37721},
  year={2026}
}

@article{ewam,
  title={EWAM: Emergent Depth-Wise Specialization in a Unified Embodied Model--From Semantic Understanding through Visual Foresight to Action},
  author={Wang, Hao and Wen, Jiajun and Liu, Jingzhi and Xue, Shuoshuo and Chen, Zhiliang and Lin, Min and Chang, Yicheng and Guo, Xiaoyu and Zhuo, Yukang and Chong, Zheng and others},
  journal={arXiv preprint arXiv:2609.39973},
  year={2026}
}

\clearpage
\appendix
\section{Additional Results}
This section provides additional experimental results to complement the evaluations in the main paper. These results offer a more detailed view of the downstream performance of \method{} following continual pre-training and task-specific post-training, illustrating its strengths and remaining challenges across manipulation tasks and evaluation settings.

\subsection{Per-task Results on RoboTwin 2.0 Benchmark}
\begin{table*}[htbp]
    \centering
    \caption{Per-task success rates of \method{} on RoboTwin 2.0 (\%, $\uparrow$). Each task is evaluated for 100 trials per setting. Task order follows the official leaderboard~\citep{robotwin-leaderboard}: tasks 1--25 are on the left and 26--50 on the right.}
    \label{tab:robotwin_per_task}
    \resizebox{\textwidth}{!}{%
    \begin{tabular}{ccc|ccc}
\toprule
Task & Clean2Clean & Clean2Random & Task & Clean2Clean & Clean2Random \\
\midrule
Adjust Bottle & 99 & 99 & Place Can Basket & 80 & 57 \\
Beat Block Hammer & 83 & 68 & Place Cans Plasticbox & 96 & 40 \\
Blocks Ranking RGB & 94 & 90 & Place Container Plate & 100 & 98 \\
Blocks Ranking Size & 68 & 54 & Place Dual Shoes & 61 & 41 \\
Click Alarmclock & 100 & 96 & Place Empty Cup & 99 & 94 \\
Click Bell & 100 & 93 & Place Fan & 75 & 66 \\
Dump Bin Bigbin & 97 & 85 & Place Mouse Pad & 70 & 53 \\
Grab Roller & 100 & 100 & Place Object Basket & 81 & 72 \\
Handover Block & 91 & 38 & Place Object Scale & 90 & 71 \\
Handover Mic & 100 & 94 & Place Object Stand & 93 & 80 \\
Hanging Mug & 41 & 31 & Place Phone Stand & 88 & 84 \\
Lift Pot & 100 & 96 & Place Shoe & 89 & 78 \\
Move Can Pot & 93 & 62 & Press Stapler & 85 & 65 \\
Move Pillbottle Pad & 94 & 67 & Put Bottles Dustbin & 71 & 50 \\
Move Playingcard Away & 100 & 95 & Put Object Cabinet & 73 & 66 \\
Move Stapler Pad & 33 & 41 & Rotate QRcode & 89 & 77 \\
Open Laptop & 96 & 94 & Scan Object & 74 & 57 \\
Open Microwave & 96 & 59 & Shake Bottle Horizontally & 100 & 99 \\
Pick Diverse Bottles & 80 & 50 & Shake Bottle & 100 & 100 \\
Pick Dual Bottles & 97 & 55 & Stack Blocks Three & 41 & 27 \\
Place A2B Left & 97 & 89 & Stack Blocks Two & 100 & 86 \\
Place A2B Right & 92 & 80 & Stack Bowls Three & 83 & 62 \\
Place Bread Basket & 87 & 79 & Stack Bowls Two & 97 & 96 \\
Place Bread Skillet & 89 & 73 & Stamp Seal & 77 & 52 \\
Place Burger Fries & 98 & 96 & Turn Switch & 61 & 62 \\
\midrule
\rowcolor{black!6} \multicolumn{4}{c}{Mean over all 50 tasks} & 85.96 & 72.34 \\
\bottomrule
    \end{tabular}}%
\end{table*}

\Cref{tab:robotwin_per_task} reports the success rates of \method{} on all 50 RoboTwin 2.0 tasks. The policy is post-trained on clean demonstrations and evaluated for 100 trials per task in the Clean2Clean and Clean2Random settings, with the latter introducing domain randomization. Performance remains strong in both settings on several tasks: Grab Roller and Shake Bottle each achieve 100\% success, while Adjust Bottle reaches 99\%. Other tasks show larger gaps between the two settings. For example, success rates decrease from 96\% to 40\% on Place Cans Plasticbox and from 91\% to 38\% on Handover Block. These differences reveal task-dependent sensitivity to scene variation: strong performance under clean conditions does not always translate into equally reliable execution under randomization.

\subsection{Per-task Results on RoboCasa-Panda Benchmark}
\begin{table*}[!htbp]
    \centering
    \caption{Per-task success rates of \method{} on RoboCasa-Panda (\%, $\uparrow$). Each value is the arithmetic mean over three evaluation seeds (195, 196, and 197), with 50 trials per task and seed (150 trials per task; 3,600 trials in total). Tasks follow the evaluation-protocol order, with tasks 1--12 on the left and 13--24 on the right.}
    \label{tab:robocasa_per_task}
    \begin{tabular}{cc|cc}
\toprule
Task & Mean SR (\%) $\uparrow$ & Task & Mean SR (\%) $\uparrow$ \\
\midrule
PnPCounterToCab & 54.7 & OpenDrawer & 84.0 \\
PnPCabToCounter & 21.3 & CloseDrawer & 97.3 \\
PnPCounterToSink & 68.0 & TurnOnStove & 49.3 \\
PnPSinkToCounter & 58.0 & TurnOffStove & 22.7 \\
PnPCounterToMicrowave & 41.3 & TurnOnSinkFaucet & 72.0 \\
PnPMicrowaveToCounter & 34.7 & TurnOffSinkFaucet & 94.0 \\
PnPCounterToStove & 75.3 & TurnSinkSpout & 82.0 \\
PnPStoveToCounter & 67.3 & CoffeeSetupMug & 25.3 \\
OpenSingleDoor & 77.3 & CoffeeServeMug & 61.3 \\
CloseSingleDoor & 92.0 & CoffeePressButton & 85.3 \\
OpenDoubleDoor & 95.3 & TurnOnMicrowave & 93.3 \\
CloseDoubleDoor & 90.0 & TurnOffMicrowave & 100.0 \\
\midrule
\rowcolor{black!6} \multicolumn{3}{c}{Mean over all 24 tasks} & 68.4 \\
\bottomrule
    \end{tabular}
\end{table*}

\Cref{tab:robocasa_per_task} presents the per-task success rates of \method{} on the 24 RoboCasa-Panda tasks. Each value is averaged over three evaluation seeds, with 50 trials per task and seed. The breakdown reveals substantial variation across kitchen manipulation tasks. \method{} performs well on several articulated-object interactions, reaching 97.3\% on CloseDrawer and 95.3\% on OpenDoubleDoor. Strong performance also extends to appliance operation, with 93.3\% success on TurnOnMicrowave.

\subsection{Per-task Results on RoboDojo-Real Benchmark}
\begin{table*}[!htbp]
    \centering
    \caption{Per-task results of \method{} on the RoboDojo real-world benchmark. Score ($\uparrow$) is scaled to 0--100, and SR (\%, $\uparrow$) measures task completion. Means give equal weight to the six tasks within each embodiment; the overall average gives equal weight to the three embodiments.}
    \label{tab:robodojo_per_task}
    \begin{tabular}{cccc}
\toprule
Embodiment & Task & Score $\uparrow$ & SR (\%) $\uparrow$ \\
\midrule
\multirow{6}{*}{ARX X5} & cover\_blocks & 65.0 & 50 \\
 & store\_in\_safe & 100.0 & 100 \\
 & pack\_and\_pour\_fruit & 70.0 & 40 \\
 & make\_bread & 0.0 & 0 \\
 & make\_food & 29.0 & 20 \\
 & insert\_tubes & 40.0 & 20 \\
\rowcolor{black!6}  & Mean & 50.7 & 38.3 \\
\midrule
\multirow{6}{*}{Piper} & stand\_up\_bottles & 95.0 & 90 \\
 & stack\_bowls & 90.0 & 90 \\
 & stack\_and\_cover\_blocks & 30.0 & 10 \\
 & insert\_charger & 5.0 & 0 \\
 & fill\_pen\_holder & 72.0 & 60 \\
 & put\_objects\_into\_basket & 70.0 & 70 \\
\rowcolor{black!6}  & Mean & 60.3 & 53.3 \\
\midrule
\multirow{6}{*}{Piper X} & sweep\_blocks & 0.0 & 0 \\
 & disassemble\_LEGO & 0.0 & 0 \\
 & classify\_objects & 48.0 & 40 \\
 & hang\_mugs & 8.0 & 0 \\
 & cap\_pen & 18.0 & 0 \\
 & pack\_objects\_into\_backpack & 22.0 & 20 \\
\rowcolor{black!6}  & Mean & 16.0 & 10.0 \\
\midrule
\rowcolor{black!6} \multicolumn{2}{c}{Overall average} & 42.3 & 33.9 \\
\bottomrule
    \end{tabular}
\end{table*}

\Cref{tab:robodojo_per_task} details the results of \method{} on 18 RoboDojo-Real tasks across ARX X5, Piper, and Piper X, following joint post-training of a single policy on all three embodiments. Scores on a 0--100 scale capture partial task progress, whereas success rates measure complete task execution. On ARX X5, store\_in\_safe achieves a score of 100.0 and a success rate of 100\%. On Piper, stand\_up\_bottles and stack\_bowls each reach 90\% success. On Piper X, classify\_objects achieves 40\% success, while cap\_pen and hang\_mugs obtain scores of 18.0 and 8.0 despite zero complete successes.



\subsection{RoboDojo-Sim Benchmark}
\label{sec:robodojo_sim}
\begin{table*}[!htbp]
    \centering
    \caption{\textbf{RoboDojo simulation benchmark results across five capability dimensions.} Score ($\uparrow$) captures partial task progress, and SR (\%, $\uparrow$) measures binary task completion. Results of the compared methods are taken from the official leaderboard~\citep{robodojo-leaderboard}, and \method{} is evaluated locally. Per-task results are shown in \Cref{tab:robodojo_sim_per_task}. \textbf{Bold} and \underline{underline} denote the best and second-best displayed values.}
    \label{tab:robodojo_sim}
    \setlength{\tabcolsep}{4pt}
    \resizebox{\textwidth}{!}{%
    \begin{tabular}{lcccccccccccc}
\toprule
\multirow{2}{*}{Method} & \multicolumn{2}{c}{Generalization} & \multicolumn{2}{c}{Precision} & \multicolumn{2}{c}{Long-Horizon} & \multicolumn{2}{c}{Memory} & \multicolumn{2}{c}{Open} & \multicolumn{2}{c}{Average} \\
\cmidrule(lr){2-3}\cmidrule(lr){4-5}\cmidrule(lr){6-7}\cmidrule(lr){8-9}\cmidrule(lr){10-11}\cmidrule(l){12-13}
 & Score & SR & Score & SR & Score & SR & Score & SR & Score & SR & Score & SR \\
\midrule
InternVLA-A1.5~\citep{internvla-a1} & 10.4 & 6.8 & 15.2 & 10.2 & 23.8 & 13.8 & 4.9 & 3.6 & 1.4 & 1.4 & 11.2 & 7.1 \\
$\pi_{0.5}$~\citep{pi05} & 13.4 & 8.2 & 12.4 & 5.5 & 23.5 & 14.7 & 5.9 & 4.7 & 2.0 & 1.7 & 11.4 & 6.9 \\
SimpleMemVLA~\citep{simplememvla} & 6.4 & 4.0 & 7.4 & 2.9 & 14.6 & 5.5 & \textbf{33.7} & \textbf{33.2} & 0.9 & 0.8 & 12.6 & 9.3 \\
KinRT~\citep{kinrt} & 14.0 & 8.6 & 15.7 & 9.9 & 26.4 & \underline{18.1} & 4.8 & 3.6 & \textbf{4.2} & \textbf{3.8} & 13.0 & 8.8 \\
CogWAM~\citep{cogwam} & 12.8 & 9.3 & \underline{20.5} & \textbf{14.6} & \underline{27.4} & \textbf{18.2} & 8.2 & 7.0 & \underline{2.9} & \underline{2.7} & 14.3 & \underline{10.3} \\
EWAM~\citep{ewam} & \underline{15.1} & \textbf{9.7} & \textbf{22.2} & \underline{12.3} & \textbf{30.2} & \underline{18.1} & 7.4 & 6.0 & 1.1 & 0.8 & \underline{15.2} & 9.4 \\
\midrule
\rowcolor{black!6} Spatial Forcing & 14.1 & 9.3 & 17.3 & 10.6 & 23.3 & 14.6 & 5.4 & 4.1 & 1.8 & 1.6 & 12.4 & 8.0 \\
\rowcolor{black!6} \method{} (Ours) & \textbf{15.3} & \underline{9.4} & 16.0 & 11.2 & 22.3 & 11.9 & \underline{26.2} & \underline{25.8} & 2.0 & 1.9 & \textbf{16.4} & \textbf{12.0} \\
\bottomrule
    \end{tabular}}%
\end{table*}
\begin{table*}[!htbp]
    \centering
    \caption{Per-task results of \method{} on the RoboDojo simulation benchmark. Score ($\uparrow$) is scaled to 0--100, and SR (\%, $\uparrow$) measures task completion, as in \Cref{tab:robodojo_sim}. Task order follows the official leaderboard~\citep{robodojo-leaderboard}, with each Generalization task followed by its randomized (\_random) variant; configurations 1--27 are on the left and 28--54 on the right. The overall average gives equal weight to the five capability dimensions.}
    \label{tab:robodojo_sim_per_task}
    \small
    \setlength{\tabcolsep}{4pt}
    \begin{tabular}{ccc|ccc}
\toprule
Task & Score $\uparrow$ & SR (\%) $\uparrow$ & Task & Score $\uparrow$ & SR (\%) $\uparrow$ \\
\midrule
stack\_bowls & 74.5 & 70 & pour\_balls\_into\_vase & 20.0 & 20 \\
stack\_bowls\_random & 19.0 & 10 & play\_Xylophone & 0.0 & 0 \\
push\_T & 0.0 & 0 & deposit\_coin & 2.0 & 0 \\
push\_T\_random & 0.0 & 0 & insert\_key & 12.8 & 0 \\
pack\_objects\_into\_box & 17.2 & 5 & build\_tower & 69.0 & 65 \\
pack\_objects\_into\_box\_random & 14.8 & 0 & put\_bottles\_into\_dustbin & 76.8 & 65 \\
fold\_clothes & 39.0 & 35 & play\_tic\_tac\_toe & 3.0 & 0 \\
fold\_clothes\_random & 29.0 & 20 & classify\_objects & 34.0 & 25 \\
hang\_mugs & 14.8 & 0 & fill\_pen\_holder & 21.2 & 5 \\
hang\_mugs\_random & 3.0 & 0 & fill\_egg\_holder & 2.2 & 0 \\
sweep\_blocks & 5.0 & 5 & organize\_table & 41.2 & 0 \\
sweep\_blocks\_random & 0.0 & 0 & play\_stacking\_toy & 0.0 & 0 \\
pour\_liquid\_into\_cup & 45.0 & 45 & make\_kong & 0.0 & 0 \\
pour\_liquid\_into\_cup\_random & 15.0 & 15 & cover\_blocks & 92.2 & 90 \\
make\_toast & 13.8 & 0 & match\_and\_pick\_from\_conveyor & 65.0 & 65 \\
make\_toast\_random & 2.5 & 0 & swap\_T & 0.0 & 0 \\
arrange\_largest\_number & 5.5 & 0 & press\_by\_number & 0.0 & 0 \\
arrange\_largest\_number\_random & 1.0 & 0 & imitate\_sorting\_sequence & 0.2 & 0 \\
sort\_nesting\_dolls\_by\_size & 0.0 & 0 & swap\_blocks & 0.0 & 0 \\
sort\_nesting\_dolls\_by\_size\_random & 0.0 & 0 & align\_blocks & 0.0 & 0 \\
store\_laptop\_and\_headphones & 34.0 & 15 & solve\_equation & 0.0 & 0 \\
store\_laptop\_and\_headphones\_random & 15.0 & 0 & stack\_blocks\_by\_language & 0.0 & 0 \\
stack\_blocks & 17.0 & 5 & general\_pickup & 15.0 & 15 \\
stack\_blocks\_random & 1.5 & 0 & classify\_objects\_by\_language & 1.0 & 0 \\
fasten\_screws & 4.5 & 0 & pick\_from\_conveyor\_by\_image & 0.0 & 0 \\
insert\_tubes & 20.0 & 5 & store\_tools\_in\_toolbox & 0.0 & 0 \\
plug\_in\_charger & 0.0 & 0 & pour\_by\_language & 0.0 & 0 \\
\midrule
\rowcolor{black!6} \multicolumn{4}{c}{Mean over all five dimensions} & 16.4 & 12.0 \\
\bottomrule
    \end{tabular}
\end{table*}

\Cref{tab:robodojo_sim} compares \method{} with methods from the official RoboDojo-Sim leaderboard~\citep{robodojo-leaderboard}, which evaluates 42 base tasks across five capability dimensions; each Generalization task also has a randomized variant, giving 54 task configurations. A single policy is post-trained on the simulation demonstrations with the memory mechanism in \cref{sec:memory}; for this benchmark, spatial alignment is applied only during continual pre-training. \method{} obtains the highest average Score and SR among the compared methods, 16.4 and 12.0\%, exceeding $\pi_{0.5}$ by 5.0 points and 5.1 percentage points. The largest gain appears in Memory, where \method{} reaches 26.2 and 25.8\%, second only to SimpleMemVLA and well above the other methods, which reach at most 8.2 and 7.0\%. Generalization achieves the highest Score of 15.3, whereas Long-Horizon remains below most baselines at 22.3 and 11.9\%.

\Cref{tab:robodojo_sim_per_task} details the results on all 54 configurations. The best result is obtained on cover\_blocks, with a Score of 92.2 and 90\% success, while stack\_bowls, put\_bottles\_into\_dustbin, build\_tower, and match\_and\_pick\_from\_conveyor reach 65\% to 70\% success. Within Memory, however, swap\_blocks, swap\_T, press\_by\_number, and imitate\_sorting\_sequence have no successful episodes. Randomized layouts substantially reduce performance: success decreases from 70\% to 10\% on stack\_bowls and from 45\% to 15\% on pour\_liquid\_into\_cup. Several tasks also show partial progress without completion, such as organize\_table with a Score of 41.2 and 0\% success. In the Open dimension, general\_pickup is the only task with non-zero success, and seven of the eight Open tasks are absent from the post-training demonstrations. These results indicate that randomized layouts, full multi-stage completion, and tasks without demonstrations remain the main challenges.

\section{Memory Mechanism}
\label{sec:memory}

As illustrated in \Cref{sec:dojo-real}, we utilize a memory mechanism for RoboDojo-Real post-training stage. Several recent VLA models incorporate memory through dedicated modules~\citep{shi2025memoryvla,shi2026memoryvla++,zeng2026kemo,sridhar2025memer,li2026remem}. Our memory mechanism instead supplies selected past frames of the main camera to the policy as additional image inputs, which are processed by the same vision-language backbone as the current views.

\subsection{Memory Construction}
Let $I_k$ denote the $k$-th main-camera frame of the current episode, where $k=0$ is the first frame, and let $f$ be the recording frequency of the embodiment ($f=25$ for Piper X and $f=30$ for ARX X5 and Piper). When generating an action chunk at frame $t$, the memory context consists of nine main-camera frames:
\begin{equation}
    \mathcal M_t=\big[\,I_0,\; I_{\max(t-8f,\,0)},\; I_{\max(t-7f,\,0)},\; \ldots,\; I_{\max(t-f,\,0)}\,\big],
    \label{eq:memory_context}
\end{equation}
that is, the episode-initial frame followed by one frame per second over the preceding eight seconds, ordered from the oldest to the most recent. The initial frame records the scene layout before manipulation and remains available throughout the episode, while the recent frames capture short-term progress and motion that a single frame cannot convey. Within the first eight seconds of an episode, the missing recent frames are replaced by the initial frame. Memory frames are taken only from the main camera, while the wrist cameras contribute only their current frames.

\subsection{Memory Encoding}
Each memory frame is resized to the resolution of the current views and encoded by the same SigLIP encoder~\citep{siglip} into 256 visual tokens. Within the prefix, the image tokens are arranged in temporal order, namely the initial frame, the recent frames from the oldest to the most recent, and the three current views, followed by the language tokens. This order is conveyed only by the token positions, without additional temporal embeddings. Like the current-view tokens, the memory tokens interact with the other prefix tokens through full attention and are accessible to the action expert. The memory frames are excluded from the spatial alignment branch: they are not passed to OmniVGGT, and $\mathcal L_{\mathrm{align}}$ in \Cref{eq:align_loss} is computed only over the current-view tokens.

\subsection{Training and Inference}
During post-training, the data loader retrieves the memory frames of each training sample according to \Cref{eq:memory_context}.During deployment, an online buffer records the main-camera frame at the camera frequency, including the control steps at which no new action chunk is requested, and retains the initial frame together with the frames of the most recent eight seconds. Whenever a new action chunk is requested, the buffer assembles $\mathcal M_t$ following \Cref{eq:memory_context}, so that the memory observed at inference matches the one constructed during training. The buffer is cleared at the start of each episode to prevent leakage across episodes.

\end{document}